\documentclass{article}

\makeatletter
\usepackage{yeyt}

\usepackage{hyperref}
\usepackage{authblk}
\usepackage[noend]{algpseudocode}

\usepackage{stmaryrd}
\usepackage[flushleft]{threeparttable}
\usepackage[bottom]{footmisc}

\newcommand{\R}{{\mathbb R}}

\renewcommand{\bs}{{\mathbf s}}
\renewcommand{\bz}{{\mathbf z}}
\renewcommand{\bx}{{\mathbf x}}
\renewcommand{\by}{{\mathbf y}}
\renewcommand{\bw}{{\mathbf w}}
\renewcommand{\bh}{{\mathbf h}}
\renewcommand{\bg}{{\mathbf g}}
\renewcommand{\btau}{\pmb{\tau}}
\newcommand{\bff}{{\mathbf f}}
\newcommand{\bdelta}{\pmb{\delta}}
\renewcommand{\bsigma}{\pmb{\sigma}}
\newcommand{\beps}{\pmb{\epsilon}}
\renewcommand{\bmu}{\pmb{\mathbf \mu}}
\renewcommand{\bnu}{\pmb{\mathbf \nu}}
\newcommand{\hbmu}{\hat{\pmb{\mu}}}
\newcommand{\hbsigma}{\hat{\pmb{\sigma}}}

\newcommand{\model}{{\em BEM}}

\newcommand{\E}{{\mathbb E}}

\newcommand{\X}{\mathcal{X}}

\newcommand{\Prob}{\mathbb{P}}

\begin{document}
\title{Bayes EMbedding (BEM): Refining Representation by Integrating Knowledge Graphs and Behavior-specific Networks}

\author[1]{Yuting Ye\thanks{yeyt@berkeley.edu}}
\author[2]{Xuwu Wang\thanks{18110240012@fudan.edu.cn}}
\author[3]{Jiangchao Yao\thanks{jiangchao.yjc@alibaba-inc.com}}
\author[3]{Kunyang Jia\thanks{kunyang.jky@alibaba-inc.com}}
\author[3]{Jingren Zhou\thanks{jingren.zhou@alibaba-inc.com}}
\author[2]{Yanghua Xiao\thanks{shawyh@fudan.edu.cn}}
\author[3]{Hongxia Yang\thanks{yang.yhx@alibaba-inc.com}}
\affil[1]{Division of Biostatistics, University of California, Berkeley, U.S.}
\affil[2]{School of Computer Science, Fudan University, China}
\affil[3]{Alibaba Group, China}

\date{\today} 

\maketitle

\begin{abstract}
Low-dimensional embeddings of knowledge graphs and behavior graphs %% (defined by context-specific relations; e.g., for the item-click-item graph in the e-commerce area, two items are linked if they are clicked by the same user)
have proved remarkably powerful in varieties of tasks, from predicting unobserved edges between entities to content recommendation. The two types of graphs can contain distinct and complementary information for the same entities/nodes. However, previous works focus either on knowledge graph embedding or behavior graph embedding while few works consider both in a unified way. 

Here we present \model\/, a Bayesian framework that incorporates the information from knowledge graphs and behavior graphs. To be more specific, \model\ takes as prior the pre-trained embeddings from the knowledge graph, and integrates them with the pre-trained embeddings from the behavior graphs via a Bayesian generative model. \model\ is able to mutually refine the embeddings from both sides while preserving their own topological structures. To show the superiority of our method, we conduct a range of experiments on three benchmark datasets: node classification, link prediction, triplet classification on two small datasets related to Freebase, and item recommendation on a large-scale e-commerce dataset.
\end{abstract}
\textbf{Keywords: }Knowledge Graph, Bayesian Model, Graph Embedding.

%\setcounter{tocdepth}{1}
%\tableofcontents
%-------------------------------------------------------------------------------------------------
%-------------------------------------------------------------------------------------------------

%%%%%%%%%%%%%%%%%%
%% Introduction %%
%%%%%%%%%%%%%%%%%%
\section{Introduction}\label{sec:intro}
Graphs widely exist in the real world, including social networks \cite{GraphSage:HamiltonYL17,kipf2017semi}, physical systems \cite{Battaglia:2016,Sanche:2018}, protein-protein interaction networks \cite{Fout:2017}, knowledge graphs \cite{Hamaguchi:2017} and many other areas \cite{Elias:2017}. %% One challenging but interesting fact is that,
There may be different views of the same set of nodes, and thus graphs of different architectures are built. For example, in the e-commerce industry, item-item networks can be constructed based on the user behaviors of clicks, purchases, add-to-preferences and add-to-carts respectively --- two items are linked if they are clicked (or via other operations) by the same user. A corresponding knowledge graph can be crafted to represent a collection of interlinked descriptions of the items, e.g., color, materials, functions. Throughout this article, we refer to the graph with respect to a certain behavior/context as the behavior graph (BG)\footnote{The concept of BG covers a wide range of conventional graphs and networks: pagelink networks (the link behavior), author-citation networks (the citation behavior), item-item interaction behavior (the co-click, co-purchase behaviors and etc.), to name a few.}, in order to distinguish it from the knowledge graph (KG) that consists of structured symbolic knowledge (triplets). KG and BG both reflect the interactions between entities/nodes in reality, but they differ in two aspects: 1) the graph structures; 2) the contained information; see Section \ref{sec:discussion_KG_BG} for a detailed discussion. The connection and the distinction between KG and BG imply that they can be complementary to each other. It is of great interest to integrate these two types of graphs in a unified way. 

\textbf{Benefits of the integration of KG and BG.} In the sequel, we give three perspectives with examples in the e-commerce industry to illustrate the benefits of incorporating KG and BG. First, KG-aided BG can achieve accurate recommendations. For instance, given a formal dress and high-heel shoes, methods based on BG alone may recommend arbitrary lipsticks. With information from the KG, it can make a better recommendation of formal lipsticks instead of sweet lipsticks, as KG has the knowledge that the dress and the shoes are associated with formal occasions. Second, KG-aided BG can do more than BG alone. Suppose a user buys a ticket to Alaska in January, the knowledge ``enjoying aurora in Alaska in winter'' is triggered in KG. So it can recommend down jacket, outdoor shoes and tripods for the aurora viewing in a freezing environment. But methods using BG-only embeddings can hardly connect the flight ticket to such outfits. Third, novel knowledge can be discovered from BG on top of the known. For example, recent clothing fashions can be inferred by the frequently co-clicked or co-purchased clothes. Then humans' common sense or other experts' knowledge can be used to identify the most likely choice of the fashion of this year. %Third, KG information can help deal with the cold start issue. To be more specific, it is possible to learn BG embeddings with KG when the BG network is very sparse. When there is not enough information of user behavior, the attributes, categories, and scenes of the products in KG will become an important ingredient for recommendation.

\textbf{Motivation.} To deal with multiple graphs, a standard practice is to embed the nodes as vectors while simultaneously integrating the information from all the sources \cite{yu2011bayesian,kloft2011local,zhai2012multiview}. To the best of our knowledge, however, there is no existing method that jointly learns the BG embedding and the KG embedding. As an alternative solution, it is common to take the pre-trained KG/BG embeddings as the input to learn the representation of BG/KG \cite{wu2016knowledge,xie2016representation,GraphSage:HamiltonYL17}. Or one can simply learn the embeddings of KG and BG separately, then incorporate them via an aggregation method, e.g., concatenation, linear combination. %% Despite the feasibility, the two strategies of using the pre-trained embeddings suffer from the loss of topological structures (node interactions).
For the first strategy, the interaction information contained in the KG/BG embedding can be distorted if it does not agree with that of BG/KG. For the second strategy, the topological structure from either side is either disguised (e.g., concatenating a short embedding with a long embedding) or destroyed (e.g., taking the average of two embeddings of the same length). In this article, we work with the pre-trained BG and KG embeddings as this strategy is widely applicable. Our goal is to integrate BG and KG without losing the topological information from both sides.

\begin{figure}[ht]
  \centering
  \includegraphics[width=\textwidth]{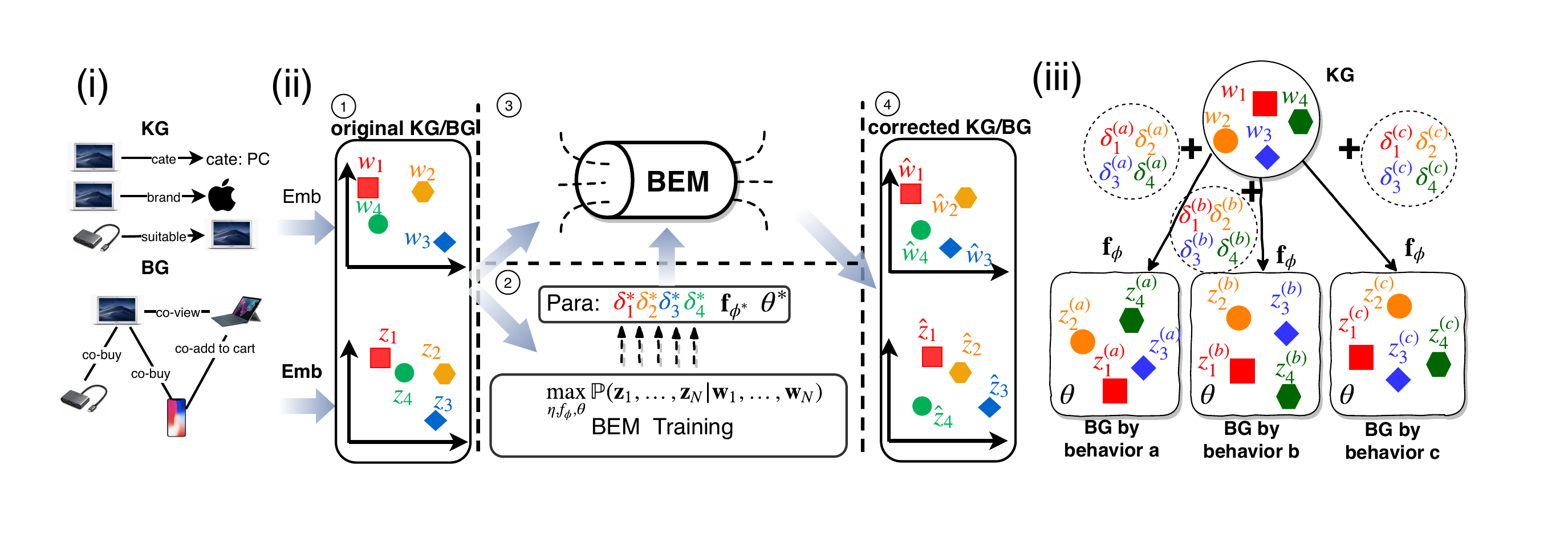}
  \vspace{-1cm}
  \caption{(i) Examples of KG and BG; (ii) The workflow of \model\/: 1) Embed KG/BG; 2) Train BEM with the parameters of the generative model in (iii); 3) Feed the original embeddings and trained parameters into BEM for refining; 4) Refined (corrected) KG/BG embeddings. (iii) A top-down generative model (Equation \eqref{eq:connection-w-z-general}) that connects one KG and three BGs with  different behaviors \textbf{a}, \textbf{b}, \textbf{c}. First, for each behavior, there exists a behavior-specific correction term $\bdelta$ that accounts for the associative bias. Then the refined KG embedding is projected into the BG space via a non-linear transformation function $\bff_{\phi}$. Finally, the BG embeddings are sampled from a distribution $p_{\theta}$ given the projected KG embedding. The model is trained to find the optimal $(\bdelta^*, \bff_{\phi^*}, \theta^*)$ that maximizes the likelihood of observing the BG embeddings $\bz$'s given the KG embeddings $\bw$'s.}
  \label{fig:BEM_illustration}
\end{figure}

\textbf{Contribution.} Throughout this paper, we consider only one KG and one BG. We develop a Bayesian framework called \model\ (Bayes EMbedding) that refines the KG and BG embeddings in an integrated fashion while preserving and revealing the topological information from the two sources. The key idea behind \model\ is that the KG embedding, plus a behavior-specific bias correction term, acts as the prior information for the generation of the BG embedding; see Figure \ref{fig:BEM_illustration} (c). \model\ aims to maximize the likelihood under this Bayesian generative model. Our contribution is twofold. From the perspective of modelling, \model\ is proposed to bridge KG and BG seamlessly, with the consideration of their respective topological structures. As a framework, \model\ is general and flexible in that it can take any pre-trained KG embeddings and any BG embeddings to mutually refine themselves.

 %% We show that \model\ is theoretically sound and practically effective in the sense that it is able to preserve the topological information.
%We demonstrate the utility of \model\ in three application studies involving two small datasets related to Freebase and a large dataset in e-commerce. We test the \model\/-refined embeddings in varieties of downstream tasks. 

The rest of the paper is organized as follows. In Section \ref{sec:discussion_KG_BG}, we discuss the difference and connection between KG and BG. In Section \ref{sec:related}, we review works that are related to our method. In Section \ref{sec:methods}, we present our method \model\/. In the sequel, we demonstrate the utility of \model\ in three application studies involving two small datasets related to Freebase and a large dataset in e-commerce (Section \ref{sec:experiments}). We test the \model\/-refined embeddings in varieties of downstream tasks. 
%we evaluate our algorithm on two small datasets and one large-scale dataset in Section \ref{sec:experiments}, which tests \model\/'s ability to integrate information from BG and KG while preserving their own topological structures.
Finally, we conclude with a discussion of the \model\ framework and highlight promising directions for future work in Section \ref{sec:conclusion}.

%% \begin{itemize}
%% \item \model\ is proposed to bridge KG and BG seamlessly, with the consideration of the behavior-specific bias. This framework provides a new perspective of making a reasoning mechanism (cognitive graph).
%% \item As a method, \model\ is generic and flexible in that it can use any KG embeddings to correct any BG embeddings. On the contrary, it is potentially able to help KG embeddings acquire novel knowledge from the BG embeddings that does not exist in the knowledge graph.
%% %\item A careful analysis is conducted to provide insights into the intrinsic nature of \model\/. Detailed derivation is executed that users can easily follow to adapt \model\ to their own needs. 
%% %\item A behavior-specific term is introduced to account for the bias between knowledge graph and the corresponding initiated behavior; \yang{IS THIS CONTRIBUTION SIGNIFICANT?}
%% % \item The generative model is simplified by considering the duality between the vertex space and edge space as well as introducing the property function on edges;  \yang{THIS SENTENCE IS QUITE CONFUSING}
%% % 	\item \model\ mimics the cognitive process of human beings, thus it holds the promise of reasoning. 
%% \end{itemize}

\iffalse
Most importantly, from the results of Section \ref{sec:experiments}, we notice that \model\ can improve not only the BG embeddings on the majority of tasks, but also boost the performance of the concatenated embeddings. It implies that, the incorporation of KG is able to expose more useful information than the original ones.
\fi

\section{Discussion of KG and BG}\label{sec:discussion_KG_BG}
Here we discuss the difference and the connection between KG and BG to illustrate three points: 1) KG and BG are different and hard to jointly learn; 2) KG and BG contain complementary (distinct but related) information, and therefore it is promising to get better embeddings by integrating the two types of graphs; 3) KG and BG can be unified from two reasonable perspectives.

\textbf{Difference between KG and BG.} There are mainly two differences between KG and BG. First, KG encodes entities and their relations in the form of the triplet as $\langle h, r, t \rangle$, where $h$, $t$ and $r$ are the head entity, the tail entity and their relation. It corresponds to a directed and highly heterogeneous network. In comparison, BG is constructed based on the interplay between the nodes under certain task/behavior-specific contexts. It corresponds to an undirected network with limited number of edge types (homogeneous or less heterogeneous than KG). Figure \ref{fig:BEM_illustration} (a) shows the difference between KG and BG in terms of the network structure. The distinction in structure makes it difficult to put the two graphs in a single framework for embedding learning. Second, the triplets in KG are extracted from authentic knowledge and experience. Thus, KG is a semantic network reflecting relatively objective facts that can stand the test of time.  % For example, $\langle \text{cell~phone,~IsOfCategory,~electronics} \rangle$ is widely known and might not be falsified for a long time.
As for BG, it embodies a time-varying and behavior-biased link between nodes, which we illustrate with two examples: 1) People may buy sunglasses and swimwear at the same time in summer, but they will barely purchase these two items in winter; 2) Two sorts of sunglasses can be viewed (the click behavior) for comparison but they are rarely bought (the purchase behavior) together. The difference in information between KG and BG indicates that they can complement each other.   % suffers from an incompleteness issue (what people know is far less than what they do not know) while the BG is more self-contained (it only focuses on a special case).

\textbf{Connection between KG and BG.} Despite the distinction, KG and BG are also closely related, resembling the connection between humans' knowledge and experience.
% Human beings summarize knowledge from abundant observations. Similarly,
KG can be regarded as an abstracted graph that reflects the shared properties among multiple BGs. This bottom-up idea (from BGs to KG) implies that it is possible to acquire novel knowledge from all kinds of BGs. On the contrary, we can heuristically interpret the connection from top down, as shown in Figure \ref{fig:BEM_illustration} (c). KG contains the general information of items, e.g., the item properties (color, materials etc.), the category of the item, the concepts/scenarios\footnotemark of the category\footnotetext{Scenarios are manually crafted to include items that appear together frequently under certain conditions. For example, the sunglasses and the swimwear both belong to the scenario ``summer-beach''.}. Then, the node of BG can be thought of as being generated by adjusting the associative entity in KG with a behavior-specific correction term. For instance, the cell-phone is conceptually a portable electronics (KG). It exhibits varieties of properties under different scenarios (BG), e.g., a communication tool when connecting to others, an entertainment platform when playing games, a working/studying tool when looking up information online. The top-down idea indicates that we can use KG information to help the learning of BG. %In this article, we mainly consider the KG with a homogeneous BG, and the top-down idea is our focus.

%%%%%%%%%%%%%%%%%%%
%% Related works %%
%%%%%%%%%%%%%%%%%%%

\section{Related Work} \label{sec:related}
In this section, we review related work to our method. As to the best of our
knowledge, there is no existing method that learns the BG embeddings and the KG embeddings jointly. We first introduce multi-view learning that is closest to this goal. Then we review alternative methods, followed by classic representation learning methods for conventional graphs and knowledge graphs.

\vspace{-0.1in}
\subsection{Multi-view Embedding Learning}
In real life, entities may have different feature subsets which is called multi-view data. For instance, in e-commerce, an item may be associated with different behavior data in different scenarios, such as the data of purchases, clicks, add-to-preferences and add-to-carts. These multi-view data can be learned to get a uniform representation for one item. For this purpose, varieties of approaches have been proposed, including co-training, multiple kernel learning, and subspace learning \cite{yu2011bayesian,kloft2011local,zhai2012multiview}. In particular, many efforts have been made in multi-view network representation learning. Qu et al. \cite{qu2017attention} combines the embeddings of different network views linearly. Shi et al. \cite{shi2018mvn2vec} proposes two characteristics (preservation and collaboration), and gets node vectors by simultaneously modeling them. It is closely related to our work in the sense that it emphasizes the integration of different sources while preserving their own specialties. However, it only deals with homogeneous networks as other multi-view embedding learning methods. In contrast, our method is designed to combine BG with KG, which differ in the data structures and the contained information.

\vspace{-0.1in}
\subsection{Alternative Ways to Integrate KG and BG}
There are alternative approaches to integrate KG and BG. First, the standard practice is to embed one graph into vectors, then take the embeddings as the input of the learning for the other graph. For example, Wu et al. \cite{wu2016knowledge} embeds sequential texts, then takes them as node/entity attributes for knowledge graph learning. Xie et al. \cite{xie2016representation} learns knowledge graph embedding by using the embeddings of entity descriptions. Hamilton et al. \cite{GraphSage:HamiltonYL17} can take as input the pre-trained KG embeddings to learn BG embeddings as well. However, this line of works tends to focus on the targeted graph (the graph that uses the pre-trained embedding for learning), but the topological structures from the other graph (the graph that generates the pre-trained embedding) may be missing. Even though interaction information between nodes is contained in the pre-trained embeddings, it can be weakened or ignored if not agreeing with the topology from the targeted graph. Second, there is a even simpler strategy to integrate KG and BG, i.e., learning the embeddings of KG and BG separately, then incorporating them via an aggregation method, e.g, concatenation, linear combination \cite{GraphSage:HamiltonYL17}. Nonetheless, the topological structures from both sides are disguised or destroyed by these aggregation methods. Our work falls in the second category, and is designed to solve the above issue: it preserves and reveals the topological information when integrating BG and KG.

\vspace{-0.1in}
\subsection{Representation Learning for BG and KG}
%Graph embeddings (GEs) have become the widely applied graph analysis methods recently, due to its convincing performance and wide applicability.
Here we review methods used to pre-train BG and KG embeddings. A line of works perform graph embedding based on graph spectrum \cite{belkin2002laplacian, tang2009relational}. Some works use matrix factorization to get node embeddings \cite{yang2015network,cao2015grarep,yang2017fast}. Additionally, simple neural networks are used to generate embeddings by making the distribution of the node embeddings close to that obtained by the topological structure \cite{perozzi2014deepwalk,tang2015line}.
Recently, some graph neural network based techniques are also proposed and widely applied \cite{kipf2017semi,Ou-etal16asymmetric,GraphSage:HamiltonYL17,velickovic2017graph}. 

Since KG differs from BG due to the semantic links between entities, the above embedding methods are not applicable to KG. Many efforts have been made to embed the nodes in KG. As a seminal work, TransE \cite{bordes2013translating} learns a low dimensional vector for every entity and relation in KGs. Later extensions include TransH  \cite{Wang:2014trans}, TransR \cite{Lin:2015:LER:2886521.2886624} and STransE  \cite{DBLP:conf/naacl/NguyenSQJ16} for more flexibilities.

%%%%%%%%%%%%%
%% Methods %%
%%%%%%%%%%%%%
\section{Methods}\label{sec:methods}
\vspace{-0.05in}
\subsection{Notation}
We denote $\bw$ and $\bz$ as the KG embedding and the BG embedding with dimension $d_{\bw}$ and $d_{\bz}$ respectively. For a vector $\bx$, let $d_{\bx}$ be the dimension of $\bx$, and let $x_k$ be its $k$-th entry. We use $\odot$ for element-wise multiplication, i.e., for two vectors $\bx$ and $\by$ with length $d$, $\bx \odot \by = (x_1y_1, \ldots, x_dy_d)$. Denote $D_{KL}(p||q)$ as the Kullback-Leibler (KL) divergence between distributions $p$ and $q$ \cite{burnham2001kullback}. Other detailed notations used throughout this section are summarized in Table~\ref{tbl:notation_tbl}.
\begin{table}[ht]
  \caption{Notation Table.}
  \label{tbl:notation_tbl}
  \centering
\begin{tabular}{p{2cm}p{8cm}}
\hline
  Notation                                     & Meaning                                                                                                                                   \\ \hline
$e$                                        & Entity.\\
$\bw$                                         & KG embedding.\\
$\bz$                                          & BG embedding.\\
$d_{\bw}/d_{\bz}$                                        & Dimension of $\bw$/$\bz$.\\
$\bdelta$                                  & The behavior-specific correction term.\\
$\bff_{\phi}(\cdot)$                                & The nonlinear transformation that projects the refined (corrected) KG embedding into the BG space.\\
$\bnu$                                        &  Projection of the KG embedding onto the behavior space by $\bff_{\phi}$.\\
$\bg(\cdot, \cdot)$                                 & The edge function that characterizes the interaction between entities in the behavior space.\\  
$p_{\eta}$                            & The distribution of $\bdelta$.\\
$p_{\theta}$        & The distribution of $\bg(\bz_i, \bz_j)$.\\
  $\bh_{\Psi}(\cdot, \cdot)$                                   & The inference network.\\
  $\btau_i$/$\btau_{i,j}$ & all the latent variables for $e_i$/$(e_i, e_j)$.\\  
  \hline\vspace{-0.1in}
\end{tabular}
\end{table}
\vspace{-0.1in}
\subsection{The Generative Model}\label{sec:generative_model}
Section \ref{sec:discussion_KG_BG} sheds light on the bottom-up and top-down relations between KG and BG. KG is thought of as the abstract representation of an entity, and BG is its realization under certain context. We can view BG as  a mix of KG and a context-specific factor (an adjustment term), but usually it only reflects some aspect of KG (i.e., a projection of the mix). Such insights motivate us to connect KG and BG in a generative model as follows.

Throughout this paper, we focus on the case where each entity has one KG embedding and one BG embedding. Mathematically, suppose there are $N$ entities and each entity $e_i$ has a KG embedding $\bw_i$ and a BG embedding $\bz_i$. 
As depicted in Figure \ref{fig:BEM_illustration}, $\bw_i$ and $\bz_i$ act as priors and observations respectively. We use $\bdelta_i$ to model the adjustment effects between $\bw_i$ and $\bz_i$. In other words, $\bdelta_i$ acts as an residual to $\bz_i$ so that $\bdelta_i + \bw_i$ is sufficient to determine the marginal distribution of $\bz_i$ via a projection function $\bff_{\phi}$. The projection not only reflects the fact that BG characterizes KG partially, but is also technically required to map $\bdelta_i + \bw_i$ into the BG space. To be more specific, we assume the joint distribution of $(z_1, \ldots, z_{N})$ hinges on the following three components: 
\begin{itemize}
\item "Refined" KG embeddings $(\bw_1 + \bdelta_1, \ldots, \bw_{N} + \bdelta_{N})$, where $(\bdelta_1, \ldots, \bdelta_{N})$ are sampled from the behavior-specific distribution $p_{\eta}$;
\item The nonlinear transformation $\bff_{\phi}$ that projects the refined KG embedding into the BG space;
\item The distribution of BG embedding $p_{\theta}$.
\end{itemize}
Then, write the generative model as
\begin{equation}
  \begin{array}{l}
    (\bdelta_1, \ldots, \bdelta_{N}) \sim p_{\eta}(\cdot)\\
    \bnu_i = \bff_{\phi}(\bw_i + \bdelta_i)\\
    (\bz_1, \ldots, \bz_{N}) \sim  p_{\theta}(\cdot | \bnu_1, \ldots, \bnu_{N}).
  \end{array}
  \label{eq:connection-w-z-general}
\end{equation}
Our target is to optimize the following objective function: 
\iffalse
Then the goal of cracking System II boils down to finding the optimal  $(p_{\eta^*}, \bff_{\phi^*}, p_{\theta^*})$ that bridges $(\bw_1, \ldots, \bw_{N})$ and $(\bz_1, \ldots, \bz_{N})$ best. Formally speaking, our target is to find the optimal parameters to solve
\fi
\begin{eqnarray}
  && \max_{\eta, f_\phi, \theta} \log \Prob(\bz_1, \ldots, \bz_N | \bw_1, \ldots, \bw_N)\nonumber\\
  &=& \max_{\eta, f_\phi, \theta} \log \int p_{\theta}(\bz_1, \ldots, \bz_N| \bff_{\phi}(\bw_1 + \bdelta_1), \ldots, \bff_{\phi}(\bw_N + \bdelta_N)) \cdot\nonumber\\
  && {\color{white}............}  p_{\eta}(\bdelta_1, \ldots, \bdelta_N)d\bdelta_1\ldots\bdelta_N.
  \label{obj:general}
\end{eqnarray}
However, the objective function Equation \eqref{obj:general} under Model \eqref{eq:connection-w-z-general} is generally intractable. %% due to the high-order interactions among entities.
For the sake of computational feasibility, assumptions are needed to simplify the model:
\begin{itemize}
\item To reduce the model complexity, we assume $\bdelta_i$'s are identically independently distributed, i.e., $\bdelta_i \overset{i.i.d}{\sim} p_{\eta}(\cdot)$, where $\eta$ is shared by all the entities.
\item To retain the interaction information between entities, we come up with an edge function $\bg(\cdot, \cdot)$ that characterizes the interplay between $e_i$ and $e_j$. For example, $\bg(\bz_i, \bz_j)$ can be the similarity or the vector difference between $\bz_i$ and $\bz_j$. Then, $p_{\theta}$ is assumed to be a generative distribution for $\bg(\bz_i, \bz_j)$. 
\item To further reduce the model complexity, we assume $\bg(\bz_i, \bz_j)$'s are i.i.d sampled from $p_{\theta}(\cdot | \bg(\bnu_i, \bnu_j))$, where $\theta$ is shared for all pairs of $(e_i, e_j)$.
\end{itemize}
Then, Model \eqref{eq:connection-w-z-general} is reduced to
\begin{equation}
  \hspace{-0.08cm}
  \begin{array}{l}
    \bdelta_i \sim p_{\eta}(\cdot), \bdelta_j \sim p_{\eta}(\cdot)\\
    \bnu_i = \bff_{\phi}(\bw_i + \bdelta_i), \bnu_j = \bff_{\phi}(\bw_j + \bdelta_j)\\    
    g(\bz_i, \bz_j) \sim p_{\theta}(\cdot | g(\bnu_i, \bnu_j)),
  \end{array}
  \label{eq:connection-w-z-indpt-pair}
\end{equation}
which is visualized as Figure \ref{fig:BEM-Gaussian-trans} (a). Compared to Model \eqref{eq:connection-w-z-general}, the reduced model has a much smaller model complexity while retaining the interaction information between entities, i.e., preserving the topological structure, which is crucial for all the BG and KG embedding methods \cite{cui2018survey, cai2018comprehensive}. We call this model \textbf{\model\/-P} (``P'' denotes pairwise interactions). In comparison, we can ignore the interactions for further complexity reduction:
\begin{equation}
  \begin{array}{l}
    \bdelta_i \sim p_{\eta}(\cdot)\\
    \bnu_i = \bff_{\phi}(\bw_i + \bdelta_i)\\
    \bz_i \sim p_{\theta} \left (\cdot | \bnu_i \right ).
  \end{array}
  \label{eq:connection-w-z-indpt}
\end{equation}
In fact, Model \eqref{eq:connection-w-z-indpt} is a special case of Model \eqref{eq:connection-w-z-indpt-pair} by letting $\bg(\bx, \by) = (\bx, \by)$ and assuming $p_{\theta}(\cdot| \bx, \by) = p_{\theta}(\cdot| \bx)p_{\theta}(\cdot| \by)$. Then it becomes a model with full independence. We call this model \textbf{\model\/-I} (``I'' denotes vertex independence). Finally, for the sake of simplicity, we denote \textbf{\model\/-O} (``O'' denotes NULL) as using the original embeddings directly without applying \model\/. All these models are summarized in Table \ref{tbl:model_name}. In the sequel, we will omit the subscript $\eta$, $\phi$ and $\theta$ for simplicity if it does not brings about ambiguity.
\begin{table}[]
    \caption{Abbreviations of models and embeddings}
    \label{tbl:model_name}
    \centering
\begin{tabular}{l|l}
\hline
Abbreviation                               & Meaning                                                                                                                                                                                 \\ \hline
{\small \model\/-P} & \begin{tabular}[c]{@{}l@{}} {\small BEM with node interactions; Model \eqref{eq:connection-w-z-indpt-pair}.}\end{tabular}   \\ \hline
{\small \model\/-I} & \begin{tabular}[c]{@{}l@{}}{\small BEM with full independence; Model \eqref{eq:connection-w-z-indpt}.}\end{tabular} \\ \hline
{\small \model\/-O} & {\small Without applying BEM.}\\ \hline\hline
{\small $\mathcal{G}$-P, $\mathcal{G}$-I  ($\mathcal{G} \in$ \{BG, KG\})} & \begin{tabular}[c]{@{}l@{}} {\small The $\mathcal{G}$ embedding by \model\/-P, \model\/-I.} \end{tabular}\\\hline
{\small $\mathcal{G}$-O ($\mathcal{G} \in$ \{BG, KG\})} & \begin{tabular}[c]{@{}l@{}} {\small The original $\mathcal{G}$ embedding (by \model\/-O).}  \end{tabular}\\\hline
{\small concat-X (X $\in$ \{P, I, O\})} & \begin{tabular}[c]{@{}l@{}} {\small The concatenation of KG-X and BG-X.} \end{tabular}\\\hline
\end{tabular}
\end{table}

\vspace{-0.1in}  
\subsection{The Inference Model}\label{sec:influence_model}
Given  Equation \eqref{eq:connection-w-z-indpt-pair}, the objective function \eqref{obj:general} can be rewritten as
\begin{equation}
  \max \sum_{(i,j): i \neq j} \log \Prob(g(\bz_i, \bz_j)| (\bw_i,\bw_j)).
  \label{obj:w-z-indpt-pair}
\end{equation}
There are varieties of off-the-shelf methods to optimize Equation \eqref{obj:w-z-indpt-pair}, such as the EM \cite{neal1998view} or MCMC \cite{gilks1995markov} algorithm. But these methods usually fail due to intractability of scalability. To this end, we resort to variational inference \cite{blei2017variational}, which is very popular for large-scale scenarios or distributions with intractable integrals. Let $\btau_{i}$ be a set of all the latent variables for node $i$, and $\btau_{ij} = \btau_{i} \cup \btau_{j}$. For example, in the generative model Equation \eqref{eq:connection-w-z-indpt}, $\btau_{i} = \{\bdelta_i\}$, $\btau_{j} = \{\bdelta_j\}$ and $\btau_{ij} = \{\bdelta_i, \bdelta_j\}$. It is easy to derive that
\begin{eqnarray}
  &&\log \Prob(\bg(\bz_i, \bz_j) | (\bw_i, \bw_j)) \nonumber\\
  &\geq& \mathbb{E}_{q(\btau_{i,j} | \bz_i, \bz_j, \bw_i, \bw_j)} \log \mathbb{P}(\bg(\bz_i, \bz_j) | \btau_{i,j}, \bw_i, \bw_j) \nonumber\\
&&~~~ -D_{KL}(q(\btau_{i,j} | \bz_i, \bz_j, \bw_i, \bw_j) || p(\btau_{i,j}| \bw_i, \bw_j)), \label{eq:elbo-w-z-indpt-pair}
\end{eqnarray}
where  $q(\btau_{i,j} | \bz_i, \bz_j, \bw_i, \bw_j)$ is called the \textit{inference model} \cite{kingma2013auto}, i.e., an approximated density function to the posterior density of $\btau_{i,j}$ given $(\bz_i, \bz_j)$.  $p(\btau_{i,j}| \bw_i, \bw_j)$ is the associated prior density. Formula \eqref{eq:elbo-w-z-indpt-pair} is also called the variational lower bound or evidence lower bound (ELBO) \cite{hoffman2016elbo} for $\log \Prob(\bg(\bz_i, \bz_j) | \bw_i, \bw_j)$. The first term in the ELBO is termed as the reconstruction term that measures the goodness of the fit, while the second one is a penalty term that measures the distance between the approximated density to the prior density. Then, our goal of maximizing $\log \Prob(\bg(\bz_i, \bz_j) | \bw_i, \bw_j)$ can be relaxed to maximizing the ELBO. It is well-known that the naive Monte-Carlo gradient estimator exhibits very high variance  and is impractical when $N$ is large \cite{paisley2012variational}. Thus we will utilize particular distributions and introduce additional assumptions to further simplify the ELBO. %% To be more specific, we focus on the normal and the log-normal distribution, and use the translation edge function detailed as following to obtain the simplified ELBO. 

\begin{figure}[ht]
  \centering
  \includegraphics[width=4.1in]{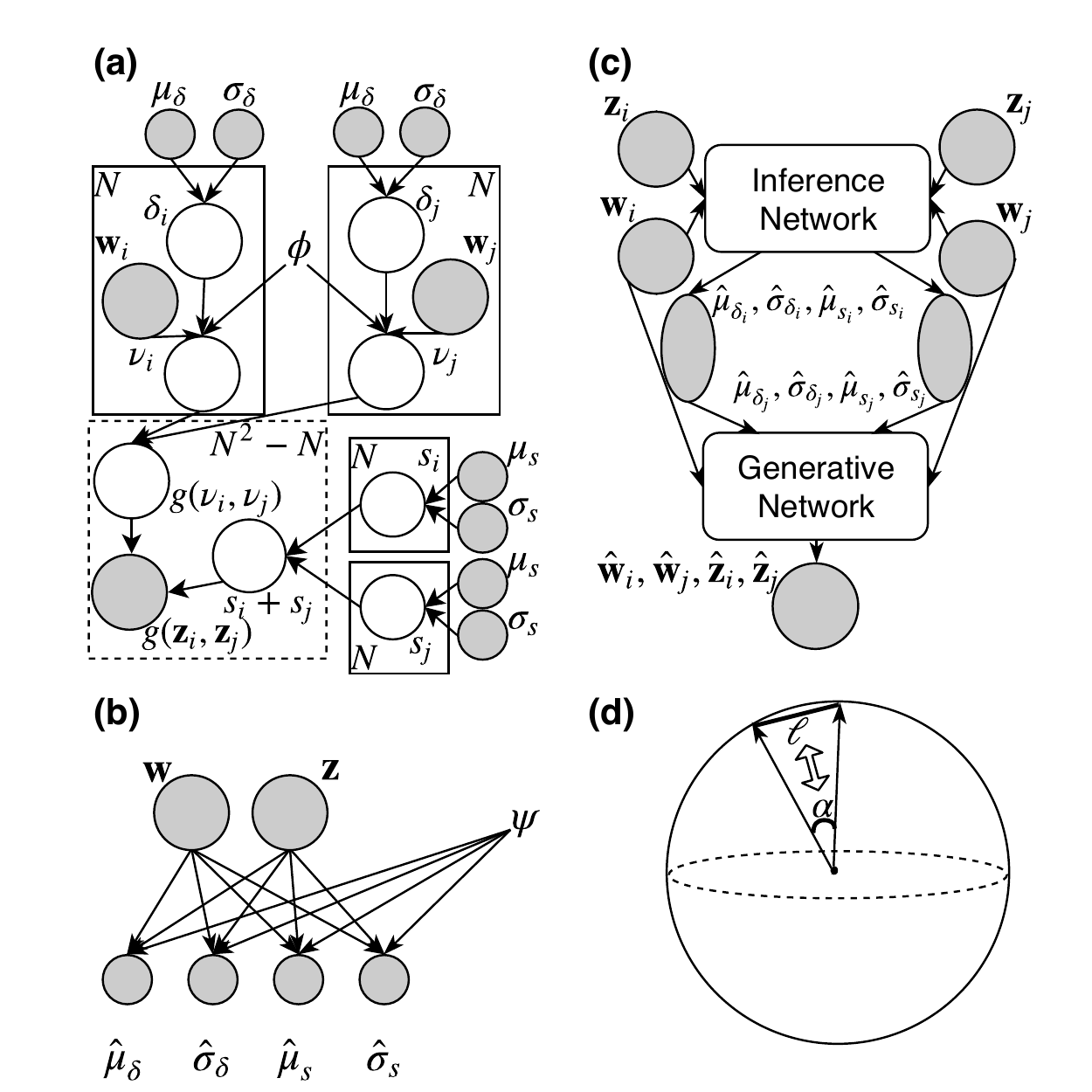}
  \caption{The \model\/-P method with the normal/log-normal distributions with the sample-specific variance $\theta_{ij} = \bs_i + \bs_j$ for $\bg(\bz_i, \bz_j)$. (a) The generative model \eqref{eq:connection-w-z-indpt-pair}. The shaded circles represent observed/estimated variables. The empty circles represent latent variables. Edges signify conditional dependency (including deterministic mapping). The solid rectangles (``plates'') indicate independent replication while the dashed rectangles indicate replication only. (b) The inference network indexed by $\psi$. It takes in KG-O and BG-O and outputs the posterior means/variances of the latent variables in (a). (c) The computational pipeline that concatenates the inference model (b) and the generative model (a) to produce refined (corrected) KG/BG embeddings. (d) The illustration graph that explains the translation edge function is equivalent to the similarity function using inner product or cosine similarity on the sphere.}  
  \label{fig:BEM-Gaussian-trans}\vspace{-0.1in}
\end{figure}

We assume $p_{\eta}(\cdot)$ to be a multivariate normal density. %% with mean $\bmu_{\bdelta}$ and variance matrix $\text{diag}(\bsigma_{\bdelta}^2)$, where $\eta=\{\bmu_{\bdelta}, \bsigma_{\bdelta}\}$.
Assume $p_{\theta}(\cdot | \bg(\bnu_i, \bnu_j)) = p_{\theta_{ij}}(\cdot | \bg(\bnu_i, \bnu_j))$ to be a multivariate normal density with mean $\bg(\bnu_i, \bnu_j)$ and variance matrix $\text{diag}(\theta_{ij})$, where $\theta_{ij} = \bs_i + \bs_j$ is the sample-specific variance (see Figure \ref{fig:BEM-Gaussian-trans} (a)). Here, $\bs_i$ and $\bs_j$ are assumed to be sampled from a multivariate log-normal distribution. %% with mean $\bmu_{s_i}$ and variance $\bsigma_{\bs_j}^2$.
We introduce the latent variable $\bs_i$ and $\bs_j$ to account for the nuisance variation induced by sampling (see Section \ref{sec:detailed_algo}). Here we choose the multivariate normal/log-normal distribution because it enjoys appealing statistical and computational properties: 1) normal/log-normal random variables are easy to sample; 2) normal/log-normal distributions can be easily reparametrized with only two parameters \cite{kingma2013auto}; 3) There is a closed-form expression for the KL divergence between two normal/log-normal distributions.

By introducing the latent variable $\bs_i$, the set of latent variables for node $j$ becomes  $\btau_{i} = \{\bdelta_i, \bs_i\}$ and $\btau_{ij} = \{\bdelta_i, \bdelta_j, \bs_i, \bs_j\}$. We then impose two common conditions in the mean-field variational inference \cite{kingma2013auto}:
\begin{itemize}
\item Both $q(\btau_{ij}|\bz_i, \bz_j, \bw_i, \bw_j)$ and $p(\btau_{ij} | \bw_i, \bw_j)$ are from mean-filed family. That is
$$
\begin{array}{l}
q(\btau_{ij}|\bz_i, \bz_j, \bw_i, \bw_j) = q(\bdelta_i| \bw_i, \bz_i) q(\bdelta_j| \bw_j, \bz_j)q(\bs_i| \bw_i, \bz_i)q(\bs_j| \bw_j, \bz_j)\\
p(\btau_{ij} | \bw_i, \bw_j) = p(\bdelta_i| \bw_i) p(\bdelta_j| \bw_j)p(\bs_i| \bw_i)p(\bs_j| \bw_j).
\end{array}
$$
\item $q(\bdelta_i | \bz_i, \bw_i)$ and $q(\bs_i | \bz_i, \bw_i)$ are normal and log-normal densities with a diagonal covariance matrix, respectively.
\end{itemize}
Thus, the approximated posterior means and variances of each element in $\btau_i$ can be represented by a function of $\bz_i$ and $\bw_i$, denoted as $\bh_{\Psi}$, which is called the \textit{inference network}. In detail,
\begin{equation}
  \bh_{\Psi}(\bz_i, \bw_i) = (\hbmu_{\bdelta_i}, \hbsigma_{\bdelta_i}, \hbmu_{s_i}, \hbsigma_{s_i}), \label{eq:posterior-est}
\end{equation}
where $\hbmu_{\bdelta_i}$, $\hbsigma_{\bdelta_i}^2$, $\hbmu_{s_i}$, $\hbsigma_{s_i}^2$ are the approximated posterior means and variances (a vector consisting of the diagonal elements of the covariance matrix) of $\bdelta_i$ and $\bs_i$ respectively. With the reparametrization trick, we can express $\bx = \hat{\bmu}_x + \hat{\bsigma}_x \odot \beps$, $\bx \in \btau_i$ and $\beps \sim N(\mathbf{0}, \mathbf{I}_{d_{\bx}})$. Correspondingly, we express their prior means and variances as $\bmu_{\bdelta_i}$, $\lambda_{\bdelta_i}\cdot \bsigma_{\bdelta_i}^2$, $\bmu_{\bs_i}$, $\lambda_{\bs_i}\cdot\bsigma_{\bs_i}^2$, where $\lambda_{\bdelta_i}$ and $\lambda_{\bs_i}$ are two tuning parameters. Then the ELBO in Equation \eqref{eq:elbo-w-z-indpt-pair} can be explicitly expressed. The reconstruction term is
\begin{eqnarray}
&&\E_{q(\btau_{i,j} | \bz_i, \bz_j, \bw_i, \bw_j)} \log \mathbb{P}(\bg(\bz_i, \bz_j) | \btau_{i,j}, \bw_i, \bw_j) \nonumber\\
&=& -\E_{\beps_{\bdelta_i}, \beps_{\bdelta_j}, \beps_{\bs_i}, \beps_{\bs_j}} \sum_{k = 1}^{d_{\bz}} \bigg\{ \frac{1}{2}\log(s_{i, k} + s_{j, k}) \nonumber \\
&&+ \frac{[\bg(\bz_{i}, \bz_{i})_k - \bg(\bff_{\phi}(\bw_i + \bdelta_i), \bff_{\phi}(\bw_j + \bdelta_j))_k]^2}{2(s_{i, k} + s_{j, k})} \bigg \} + C_0,\label{eq:recons-indpt-pair}  
\end{eqnarray}
where $C_0$ is a constant and
\begin{equation}
  \begin{array}{l}
    \bdelta_i = \hbmu_{\bdelta_i} + \hbsigma_{\bdelta_i} \odot \beps_{\bdelta_i},~~\bdelta_j = \hbmu_{\bdelta_j} + \hbsigma_{\bdelta_j} \odot \beps_{\bdelta_j},\\
    \bs_i = \hbmu_{\bs_i} + \hbsigma_{\bs_i} \odot \beps_{\bs_i},~~\bs_j = \hbmu_{\bs_j} + \hbsigma_{\bs_j} \odot \beps_{s_j}.\label{eq:reparametrization}
  \end{array}
\end{equation}
The penalty term is
\begin{eqnarray}
&&   D_{KL}(q(\btau_{i,j} | \bz_i, \bz_j, \bw_i, \bw_j) || p(\btau_{i,j}| \bw_i, \bw_j)) \nonumber\\
&=&\sum_{\bx \in \btau_{ij}}\sum_{k = 1}^{d_{\bx}} \{-\log \frac{\hat{\sigma}_{\bx, k}^2}{\lambda_{\bx} \cdot \sigma_{\bx, k}^2} +  \frac{\hat{\sigma}_{\bx, k}^2}{\lambda_{\bx} \cdot \sigma_{\bx, k}^2} + \frac{(\hat{\mu}_{\bx, k} - \mu_{\bx, k})^2}{\lambda_{\bx} \cdot \sigma_{\bx,k}^2}\} + C_1,\nonumber\\\label{eq:penalty-indpt-pair}
\end{eqnarray}
where $C_1$ is a constant.
%% \begin{eqnarray}
%%   &&\text{P}_{\psi}(\mu_x, \hbmu_x, \sigma_x, \hbsigma_x, \lambda_x)\nonumber\\
%%   &=& \sum_{k = 1}^{d_x} \{-\log \frac{\hbsigma_{x, k}^2}{\lambda_x \cdot \sigma_{x, k}^2} +  \frac{\hbsigma_{x, k}^2}{\lambda_x \cdot \sigma_{x, k}^2} + \frac{(\hbmu_{x, k} - \mu_{x, k})^2}{\lambda_x \cdot \sigma_{x,k}^2}\}.\label{eq:penalty-indpt-pair}
%% \end{eqnarray}

We can draw several implications from the closed-form expression of the ELBO. Maximizing the ELBO in Equation\eqref{eq:elbo-w-z-indpt-pair} is equivalent to minimizing the sum of Equation\eqref{eq:recons-indpt-pair} and Equation\eqref{eq:penalty-indpt-pair}, which are balanced by $\lambda_{\bdelta_i}$ and $\lambda_{\bs_i}$. Minimizing the reconstruction term forces the corrected KG/BG embeddings to behave similarly to the observed BG embeddings as per the selected edge function $\bg$. It suggests that the reconstruction term preserves the topological structure of BG.  Accordingly, minimizing Equation \eqref{eq:penalty-indpt-pair} enforces the approximated posterior mean/variance to be close to the prior mean/variance. If the prior mean of $\bdelta_i$ is set to be $\mathbf{0}$,  such minimization forces the corrected KG/BG embeddings to be close to the observed KG embeddings. It indicates the penalty term preserves the topological structure of KG. Thus, the refined KG/BG embeddings can be regarded as a mixture of information. The two parameters $\lambda_{\bdelta_i}$ and $\lambda_{\bs_i}$ act as controllers of such mixing. For example, a small $\lambda_{\bdelta_i}$ indicates the corrected embeddings squint towards the observed KG embeddings other than the observed BG embeddings, vice versa. %

\subsection{Algorithm}\label{sec:detailed_algo}
Given all the components discussed above, we can write down the detailed algorithm of \model\/. First, we sample two batches of samples of batch size $n_B$, denoted as batch $B_a$ and $B_b$; then pair them up randomly, denoted as $B_{pair} = \{(a_m, b_m)\}_{m=1}^{n_B}$. For each batch, we impose the same prior information for all the samples in this batch, and estimate 
\begin{equation}
  \begin{array}{l}
    \bmu_{\delta}^{(l)} = \mathbf{0}, l = a, b\\
    \sigma_{\bdelta, k}^{(l)} = \frac{1}{n_B - 1}\sum_{m = 1}^{n_B}(\bw_{l_m,k} - \sum_{m' = 1}^{n_B} \frac{\bw_{l_{m'}, k}}{n_B})^2,~ k = 1, \ldots, d_{\bw}, l = a, b\\
   \mu_{\bs, k}^{(a, b)} = \frac{1}{n_B} \sum_{m = 1}^{n_B} (\bg(\bz_{a_m}, \bz_{b_m})_k - \bar{\bg}^{(a, b)}_k)^2, k = 1, \ldots, d_{\bg}\\
   \sigma_{\bs, k}^{(a, b)} = \frac{1}{R} \sum_{r =1}^R (\mu_{\bs, k}^{(a, b, r)} - \bar{\mu}_{\bs, k}^{(a, b)})^2, k = 1, \ldots, d_{\bg} \text{ (bootstrap)}
  \end{array}
  \label{eq:prior-est}
\end{equation}
where $\bar{\bg}^{(a, b)} = \frac{1}{n_B} \sum_{m' = 1}^{n_B} \bg(\bz_{a_{m'}}, \bz_{b_{m'}})$, $R$ is the number of bootstrap replicates, $\bmu_{\bs}^{(a, b, r)}$ is the $r$-th bootstrap estimator of $\bmu_{\bs}^{(a, b)}$ from $B_{pair}$ ($r = 1, \ldots, R$), and $\bar{\bmu}_{\bs}^{(a, b)} = \frac{1}{R} \sum_{r = 1}^R \bmu_{\bs}^{(a, b, r)}$. Then, for each pair of sample $(a_m, b_m)$, use the inference network Equation \eqref{eq:posterior-est} to get the approximated posterior information $\hbmu_{\bdelta_{l_m}}$, $\hbsigma_{\bdelta_{l_m}}$, $\hbmu_{\bs_{l_m}}$, $\hbsigma_{\bs_{l_m}}$, $l = a, b$, $m = 1, \ldots, n_B$, as shown in Figure \ref{fig:BEM-Gaussian-trans} (b). Next, we sample $2n_B \cdot (d_{\bw} + d_{\bz})$ standard normal variables to get $\bdelta_{l_m}$ and $\bs_{l_m}$ by Equation \eqref{eq:reparametrization}, where we set $\lambda_{\bdelta_i} \equiv \lambda_1$ and $\lambda_{\bs_i} \equiv \lambda_2$, $i = 1, \ldots, N$. We obtain the ELBO in Equation \eqref{eq:elbo-w-z-indpt-pair} via Equations \eqref{eq:recons-indpt-pair}-\eqref{eq:penalty-indpt-pair}, as shown in Figure \ref{fig:BEM-Gaussian-trans} (c). Finally, We can use any optimization method, such as Adam \cite{kingma2014adam}, to update $\phi$ and $\psi$ when maximizing the ELBO. We run the above steps for $T$ times, and we can get the refined KG/BG embedding for $e_i$ by
\begin{equation}
    \hat{\bw}_i = \bw_i + \hbmu_{\bdelta_{i}}, ~~~~\hat{\bz}_i = \bff_{\phi}(\hat{\bw}_i)\label{eq:prediction}
\end{equation}

\begin{algorithm}
    \caption{The \model\ method.}
    \label{algo:BEM-full} 
  \begin{algorithmic}[1]
    \Require Pre-Trained KG/BG embeddings $(\bw_i, \bz_i)$, $i = 1, \ldots, N$; tuning parameters $\lambda_1$, $\lambda_2$; batch size $n_B$, number of iterations $T$.
  \For{$t = 1, 2, \ldots, T$}    
    \State Sample two batches $B_a$, $B_b$ of batch size $n_B$, and pair them up as $B_{pair} = \{(a_1, b_1), \ldots, (a_{n_B}, b_{n_B})\}$;
    \State Estimate the prior information by $(\bmu_{\bdelta}^{(l)}, \bsigma_{\bdelta}^{(l)}, \bmu_{\bs}^{(a,b)}, \bsigma_{\bs}^{(a,b)})$ by Equation\eqref{eq:prior-est};
    \For{$l = a, b; m = 1, \ldots, n_B$}    
      \State Get the posterior information $(\hbmu_{\bdelta_{l_m}}, \hbsigma_{\bdelta_{l_m}}, \hbmu_{\bs_{l_m}}, \hbsigma_{\bs_{l_m}})$ by Equation\eqref{eq:posterior-est};
      \State Sample a standard normal variable from $N(\mathbf{0}, \mathbf{I}_{d_{\bw}})$ and $N(\mathbf{0}, \mathbf{I}_{d_{\bz}})$ respectively. Get $\bdelta_{l_m}$ and $\bs_{l_m}$ via Equation\eqref{eq:reparametrization}.
    \EndFor
    \State Obtain the ELBO in Equation\eqref{eq:elbo-w-z-indpt-pair} via Equations \eqref{eq:recons-indpt-pair}-\eqref{eq:penalty-indpt-pair};
    \State Update $\psi$ and $\phi$ by maximizing the ELBO.
  \EndFor
  \For{$i = 1, \ldots, N$}
    \State Get the refined KB/BG embeddings $\hat{\bw}_i$ and $\hat{\bz}_i$ by Equation\eqref{eq:prediction}.
  \EndFor
  \State Denote the $\phi$ and $\psi$ in the last round as $\hat{\phi}$ and $\hat{\phi}$.
  \Ensure $\hat{\phi}$, $\hat{\psi}$, $(\hat{\bw}_i, \hat{\bz}_i), i = 1, \ldots, N$.
\end{algorithmic}
\end{algorithm}
To analyze the complexity of Algorithm \ref{algo:BEM-full}, we simply use two-layer MLPs (multi-layer perceptron) for $\bff_{\phi}$ and $\bh_{\psi}$. Let $n_h$ be the number of hidden nodes of these neural networks. Then it is easy to see that the computational complexity is $\mathcal{O}((n_z + n_w) n_h \cdot (n_{iter} + R) \cdot n_B \cdot T)$, where $n_{iter}$ is the number of iterations for the maximization step (Line $8$) in Algorithm \ref{algo:BEM-full}. If we set $T \propto N/n_B$, the computational complexity is $\mathcal{O}((n_z + n_w) n_h \cdot (n_{iter} + R) \cdot N)$. Furthermore, the storage complexity is just $\mathcal{O}(n_zn_h + n_wn_h)$, since it merely needs to keep track of two sets of parameters in $\bff_{\phi}$ and $\bh_{\psi}$. Therefore, the algorithm is efficient in both time and storage in the sense that the size of the dataset only affects the computational time linearly. However, when the dataset is too large to be entirely loaded into the CPU, the algorithm might suffer from a non-negligible overhead caused by partitioning and loading the data during the iteration. 

\subsection{Edge Function} The edge function $\bg$ in Equation \eqref{eq:connection-w-z-indpt-pair} characterizes the interplay between nodes. The choice of this function determines what kind of KG information is incorporated into the BG embeddings. We give four examples as below:
\begin{itemize}
\item A natural choice is the translation function i.e., $\bg(\mathbf{x}, \mathbf{y}) = \mathbf{x} - \mathbf{y}$, where $d_{\bg} = d_{\bw} \text{ or } d_{\bz}$. TransE and its variants are based on the translation operation, and aim to minimize the $L_2$/$L_1$ loss between the translated embedding of the head entity and the corresponding embedding of the tail entity \cite{bordes2013translating,Wang:2014trans,lin2015learning}. 
\item 
  An arbitrary similarity function can be used that measures the similarity between $\bz_i$ and $\bz_j$, where $d_{\bg} = 1$. Such choice coincides with the objective functions of the majority of BG/KG embedding methods \cite{cui2018survey, cai2018comprehensive}. For instance, GraphSAGE \cite{GraphSage:HamiltonYL17}, GCN \cite{kipf2017semi}, node2vec \cite{grover2016node2vec} etc., maximize the inner product between positive samples while minimizing this metric between negative samples.
\item If the edge function only relies on the index $i$ and $j$, such as the edge attribute between node $i$ and node $j$, \model\ becomes a supervised model.
\item If the edge function is an identity function $\bg(\mathbf{x}, \mathbf{y}) = (\mathbf{x}, \mathbf{y})$, then Model \eqref{eq:connection-w-z-indpt-pair} is reduced to Model \eqref {eq:connection-w-z-indpt}. Here, $\bg$ simply concatenates vectors $\bx$ and $\by$, thus $d_{\bg}  = 2d_{\bw} \text{ or } 2d_{\bz}$.
\end{itemize}
In this article, we use the translation function $\bg(\mathbf{x}, \mathbf{y}) = \mathbf{x} - \mathbf{y}$. In fact, the translation function is equivalent to the similarity function using inner product or cosine similarity if the embeddings are normalized onto the unit sphere, such as embeddings generated by GraphSAGE, TransE and its variants. % For those methods that do not require normalization, it is also worth considering unit-length embeddings for the improvement of performance on certain tasks\cite{levy2015improving,wilson2015controlled}.
As shown in Figure \ref{fig:BEM-Gaussian-trans} (d), the module $\ell$ of the difference between two points on the sphere is bijectively mapped to the angle $\alpha$ between the rays from the origin to the two points.

%%%%%%%%%%%%%%%%%
%% Experiments %%
%%%%%%%%%%%%%%%%%
\section{Experiments}\label{sec:experiments}
We empirically study and evaluate \model\ on two small datasets and one large-scale dataset for a variety of tasks. Each dataset consists of one KG and one BG with pre-trained node embeddings. The goal of these experiments is to show that embeddings refined by \model\ can outperform the original pre-trained embeddings on some tasks, while remaining the efficacy for most of the others:
\begin{itemize}
        \item The \textbf{node classification} task (on two small datasets) studies if \model\ can help refine the KG/BG embedding using the BG/KG embedding. It also investigates whether \model\ can reveals useful information in KG and BG for the classification purpose (Section \ref{sec:empirical_analysis}). 
        \item The \textbf{link prediction} \cite{bordes2013translating} and the \textbf{triplet classification} \cite{socher2013reasoning} (on two small datasets) investigate whether \model\ can extract useful information from BG to refine the KG embedding. 
        \item The \textbf{item recommendation} task (on the large dataset) studies whether the information in KG can enhance the performance of the BG embedding. 
\end{itemize}   
For the node classification task, we study the KG/BG and the concatenated embeddings that are refined by \model\/. In contrast, we only consider the KG embedding for the link prediction task and the triplet classification task since the two tasks are designed for the KG embedding. We only consider the BG embedding for the item recommendation for the same reasoning. %% the BG embedding since the two tasks are designed for the KG embedding only. The concatenation of the KG and BG embeddings is not suitable either, because the BG part masks the topological information contained in the KG part. To illustrate this point, we compare the result of the concat-O embedding with that of the KG embedding refined by \model\ .Here we do not consier the concatenated embedding of KG and BG since the task is designed for the BG embedding.
%% \footnote{Here we do not consider the concatenated embedding because the task is designed for the BG embedding only, and the concatenation can distort the topological structure of BG.}.

We implement\footnote{The code can be found at \url{https://github.com/Elric2718/Bayes_Embedding}.} \model\ as per Algorithm \ref{algo:BEM-full} based on \textit{tensorflow}\footnote{https://www.tensorflow.org/}. Throughout this section, we use the following default parameter setting:
\begin{itemize}
\item Functions $\bff_{\phi}$ and $\bh_{\psi}$ are implemented as two-layer MLPs with $500$ hidden nodes and the ReLU \cite{nair2010rectified} activation.
\item The batch size $n_B$ is $500$, the optimization algorithm is Adam \cite{kingma2014adam}, the learning rate is $0.001$, the number of training steps $T = N/n_B \cdot 20$.
\item $\lambda_1 = 1.0$, $\lambda_2 = 1.0$.
\end{itemize}
A discussion on the selection of the above parameters is deferred to Appendix \ref{appendix:algo_parameters}.

%\vspace{-0.1in}
\subsection{Two Small Datasets}
The two small datasets have the same KG but differ in the BGs. The shared KG is FB15K237, which is reduced from FB15K to remove the reversal relations \cite{dettmers2018convolutional}. There are $14,541$ entities, $237$ relations, and $272,115$ training triplets, $20,466$ validation triplets, $17,535$ testing triplets. The first dataset uses a pagelink network (denoted as \textit{pagelink}) that records the linkages between the wikipedia pages of entities in FB15K237. It includes $14,071$ nodes (a subset of the entities in FB15K237) and $1,065,412$ links. The second dataset comes with a short paragraph description (denoted as \textit{desc}) for each entity in FB15K237. Strictly speaking, the descriptions do not form a BG due to the lack of connection between descriptions. We regard them as an isolated graph  to evaluate \model\ under extreme conditions where BG does not contain any interplay information between nodes. See Appendix \ref{appendix:data_detail} for more details on the two datasets.

% \usepackage{multirow}
\begin{table}[]
\caption{The node classification accuracy (\%) using the refined BG/KG embeddings by \model\/. Here KG, BG and concat refer to the KG embedding, BG embedding and the concatenation of the KG and BG embeddings, respectively.}
\label{tbl:cate-classification}
\centering
\begin{tabular}{c|c| ccc || ccc}
\hline
\multicolumn{2}{c|}{\multirow{2}{*}{}}                         & \multicolumn{6}{c}{FB15K237 + \textit{pagelink}}                                                            \\ \cline{3-8} 
\multicolumn{2}{c|}{}                                          & \multicolumn{3}{c||}{node2vec}                    & \multicolumn{3}{c}{LINE}                        \\ \hline
                        & \model\ & KG             & BG             & concat         & KG             & BG             & concat         \\ \hline
\multirow{3}{*}{TransE} & O                                     & 85.59          & 75.12          & 89.39          & 85.59          & 77.57          & 89.44          \\ 
                        & I                                     & 85.51          & 82.56          & 85.97          & 86.35          & 85.44          & 87.05          \\ 
                        & P                                     & \textbf{88.89} & \textbf{86.32} & \textbf{90.29} & \textbf{88.21} & \textbf{86.27} & \textbf{90.01} \\ \hline
\multirow{3}{*}{TransD} & O                                     & 86.06          & 75.12          & 89.18          & 86.06          & 77.57          & 89.00          \\ 
                        & I                                     & 83.73          & 78.86          & 84.16          & 86.58          & 85.10          & 86.69          \\ 
                        & P                                     & \textbf{88.60} & \textbf{85.39} & \textbf{89.90} & \textbf{88.70} & \textbf{85.30} & \textbf{89.73} \\ \hline\hline
\multicolumn{2}{c|}{\multirow{2}{*}{}}                         & \multicolumn{6}{c}{FB15K237 + \textit{desc}}                                                                \\ \cline{3-8} 
\multicolumn{2}{c|}{}                                          & \multicolumn{3}{c||}{doc2vec}                     & \multicolumn{3}{c}{sentence2vec}                \\ \hline
                        & \model\ & KG             & BG             & concat         & KG             & BG             & concat         \\ \hline
\multirow{3}{*}{TransE} & O                                     & 85.32          & 75.62          & \textbf{87.92} & 85.32          & 83.42          & 88.43          \\ 
                        & I                                     & 86.19          & 81.50          & 86.41          & 87.61          & 85.18          & 88.07          \\ 
                        & P                                     & \textbf{87.68} & \textbf{81.52} & 87.86          & \textbf{88.05} & \textbf{85.82} & \textbf{88.57} \\ \hline\hline
\multirow{3}{*}{TransD} & O                                     & 85.83          & 75.62          & 88.07          & 85.83          & 83.42          & 88.52          \\ 
                        & I                                     & 86.75          & 81.44          & 86.85          & 87.96          & 84.97          & 88.07          \\ 
                        & P                                     & \textbf{87.34} & \textbf{82.24} & \textbf{88.15} & \textbf{88.36} & \textbf{86.12} & \textbf{88.86} \\ \hline
\end{tabular}
\end{table}

We use TransE \cite{bordes2013translating} and TransD \cite{ji2015knowledge} from OpenKE\footnote{https://github.com/thunlp/OpenKE \cite{han2018openke}} to pre-train KG's embeddings. Both of them are trained for $500$ epochs with dimension $d_{\bw} = 50$ and other parameters are taken as default. For the BGs, we use doc2vec \cite{le2014distributed} and sentence2vec \cite{pgj2017unsup} to pre-train \textit{desc} BG embeddings, and node2vec \cite{grover2016node2vec} and LINE \cite{tang2015line} to pre-train \textit{pagelink} BG embeddings respectively. The dimension of the BG embedding is set to be $d_{\bz} = 100$. More details on the experiment and hyper-parameter setups are included in Appendix \ref{appendix:func_detail}.

\subsubsection{Node classification}\label{sec:node_classification}
In the node classification task, there are $46$ class labels. The embeddings are fed into a multi-label logistic regression model for training and prediction. Table \ref{tbl:cate-classification} shows the results of \model\/, from which we can draw three implications. First, we observe consistent improvements of \model\/-P over \model\/-O (the original embedding) through almost all settings (accuracies boosted by $2\%$-$10\%$ for KG and BG). It indicates that we can benefit from integrating information of the two sources. Second, if the classifier is sufficiently expressive, concat-O is expected to perform the best since there is no loss of information from the input. However, concat-P turns out to perform slightly better than concat-O in most cases. It suggests that \model\/-P not only preserves the information for node classification, but also reveals signals. %% Third, KG-P performs comparably to concat-O while BG-P is inferior. This result is attributed to the fact that the original KG embedding is of better quality than the BG one. In turn, BG can gain more from the integration than KG.
Third, as we expected, \model\/-P outperforms \model\/-I since the former accounts for the pairwise interactions that are crucial for the embedding learning of KG/BG. Finally, we point out that the concatenated embedding and the KG/BG embedding are not comparable. The concatenated embedding is longer than the \model\/-refined embedding, so the classifier for the former has more parameters, thus more expressive. For a fair comparison, we study the projection of the concatenated embedding onto the BG/KG space, and the associative results are deferred to Appendix \ref{appendix:results}.

\subsubsection{Empirical analysis}\label{sec:empirical_analysis}
\begin{figure}[htbp]
\centering
\includegraphics[width = 0.8\textwidth, height = 0.4\textwidth]{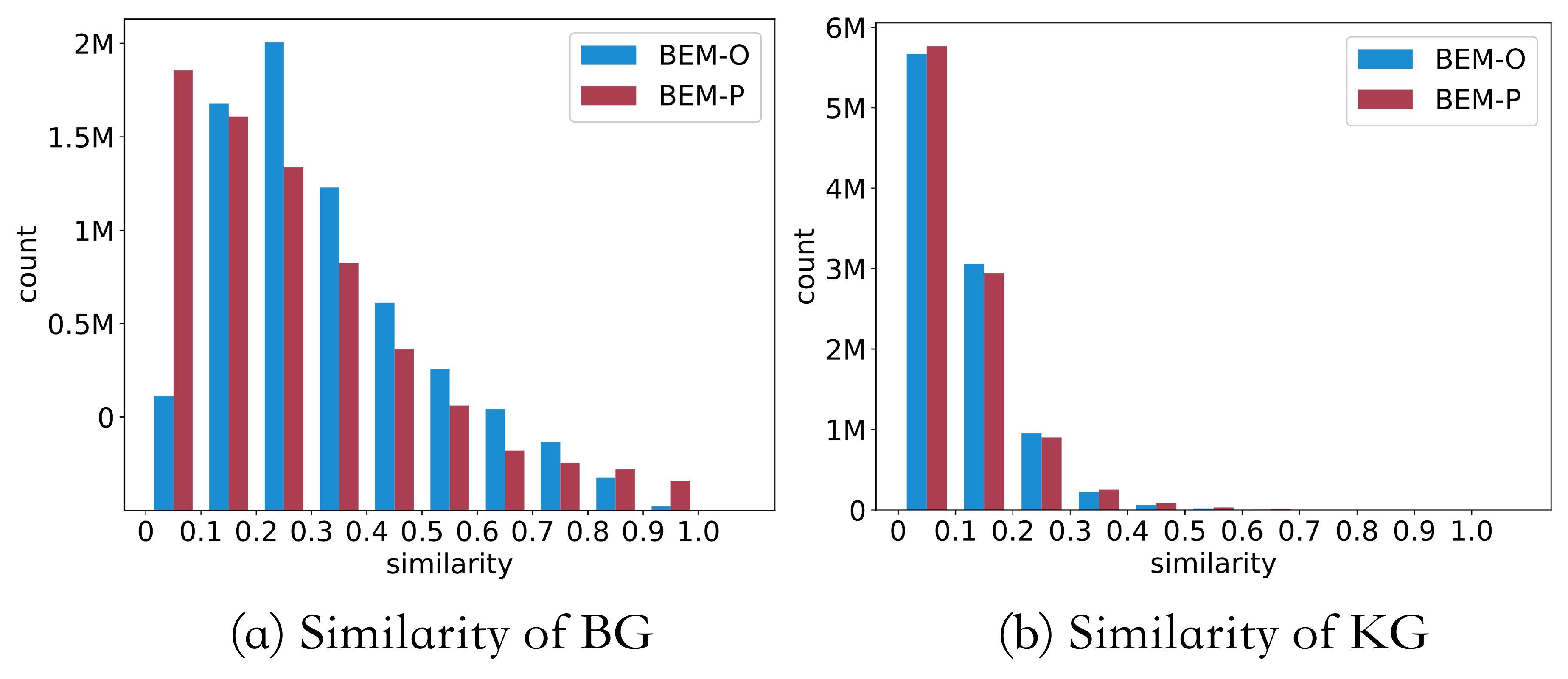}
\caption{Distribution of the similarities between nodes. Here, 1000,000 node pairs are sampled randomly.}
\label{fig:simi_raw_new}
\end{figure}

\begin{figure}[htbp]\tiny
\centering
\includegraphics[width = 0.6\textwidth, height = 0.5\textwidth]{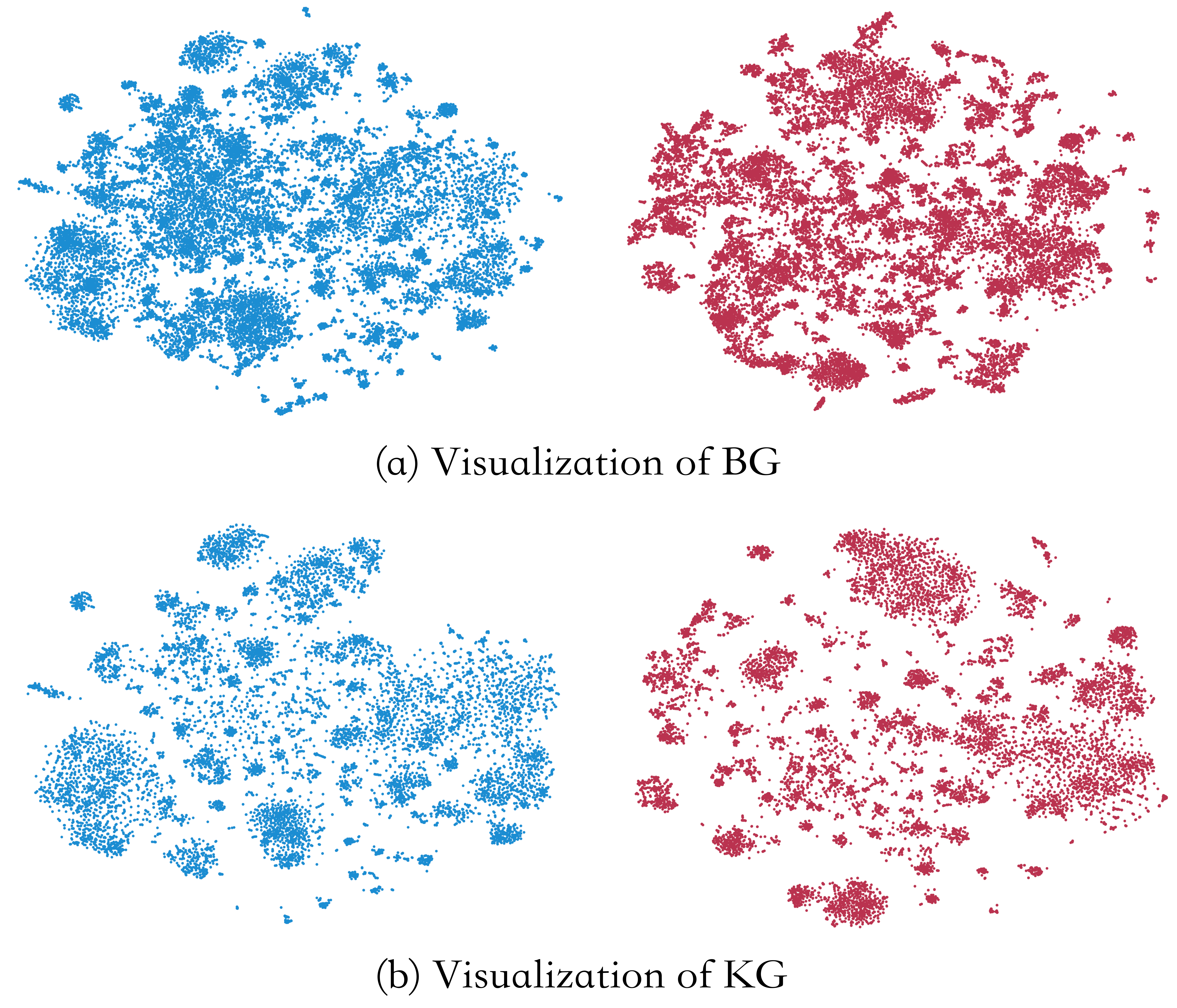}
\caption{Visualization of the embeddings. Blue: \model\/-O; Red: \model\/-P.}
\label{fig:reconstruction}
\end{figure}

To understand the property of the embeddings refined by \model\/-P, we perform two empirical data analyses on the FB15K237-\textit{pagelink} dataset. First, we compute the absolute cosine similarity for each pair of nodes using KG-O, KG-P, BG-O, BG-P respectively. From Figure~\ref{fig:simi_raw_new}, we observe that the KG-P and BG-P are distributed more extremely than KG-O and BG-O --- there are more highly correlated and more uncorrelated node pairs for the former. It indicates that \model\/-P enforces some nodes to group tightly while some others are distracted from each other. This result can also be concluded by the visualization of the embeddings using t-SNE (Figure~\ref{fig:reconstruction}). Second, we use the class labels for the node classification task to compute
$$r=\frac{\max_C \{\text{within-cluster-distance}(C)\}}{\min_{C, C'}\{\text{between-cluster-distance}(C, C')\}},$$
where $C$, $C'$ are two classes, and
$$
\begin{array}{l}
\text{within-cluster-distance}(C) = \frac{1}{|C|}\sum_{\bx \in C} ||\bx - \bar{\bx}||_2, ~~~~ \bar{\bx} = \frac{1}{|C|} \sum_{\bx \in C} \bx,\\
\text{between-cluster-distance}(C, C') = \min_{\bx \in C, \by' \in C'} ||\bx - \by'||_2.
\end{array}
$$
This metric reflects the degree to which the topological structure of the embeddings aligns with the labels. We have $r(\X_{KG-O}) = 0.3042$, $r(\X_{KG-P}) = 0.2695$, $r(\X_{BG-O}) = 0.3890$ and $r(\X_{BG-P}) = 0.3764$, indicating that \model\/-P enforces nodes in the same classes to get closer to each other while nodes across classes are pulled away. This result suggests that \model\/-P is able to preserve and further reveal the topological structure for both KG and BG. %% suggests that Theorem \ref{thm:global_local_complement} may hold with milder assumptions in reality.

\subsubsection{Link Prediction and Triplet Classification on the KG side}\label{sec:LP_TC}
We evaluate \model\ on the link prediction and the triplet classification tasks. Since \model\ can only refine the entity embeddings, we retrain the relation embedding for another $500$ epochs using \model\/-refined KG embeddings and the original relation embeddings as the initial values. In Table \ref{tbl:LP-TC}, notice that the KG embeddings can also benefit from incorporating the BG information via the \model\ refining. In contrast, the concat-O embeddings are much inferior. It validates that the concatenation does not fully expose the topological structure of KG while \model\ can make good use of this information.  %% but the gain is marginal. We speculate that it is because the BG embedding is biased towards the associative behavior. The integration of multiple BGs can mutually cancel out the biases, and find a better correction for the KG embedding. We leave this idea to future work.
%% This result is beyond our expectation because both TransE and TransD are designed to perform well in the two tasks, but the objective function of \model\ deviates from this goal.
Moreover, we observe the improvement mainly occurs for the \textit{pagelink} dataset. For the \textit{desc} dataset, the TransD embeddings get improved slightly while the TransE embeddings get worse after applying the \model\ refining. Such observation can be explained as the \textit{desc} dataset does not provide supplementary interaction information to the KG graph.

\iffalse
In addition, we have to admit that the improvements might not be significant, which is sensible as well --- a single observation comes along with behavior-specific bias and distracts the abstract knowledge that should be summarized from multiple instances. It implies that there might be an even better improvement if we have BG embeddings from distinct behaviors.
\fi

% \usepackage{multirow}
\begin{table}[]
\caption{Results of Link prediction (LP) and Triplet classification (TC).}
\label{tbl:LP-TC}
\centering
\begin{tabular}{p{2cm}|p{1.6cm}|p{1.2cm}p{0.8cm}p{1.2cm}p{0.8cm}||p{1cm}p{1.8cm}p{1cm}p{1cm}}
\hline
\multirow{3}{*}{Metrics}                                                                  & \multirow{3}{*}{Embedding} & \multicolumn{4}{c||}{FB15K237 + \textit{pagelink}}                          & \multicolumn{4}{c}{FB15K237 + \textit{desc}}                              \\ \cline{3-10} 
                                                                                          &                            & \multicolumn{2}{c}{TransE}     & \multicolumn{2}{c||}{TransD}     & \multicolumn{2}{c}{TransE}     & \multicolumn{2}{c}{TransD}     \\ \cline{3-10} 
                                                                                          &                            & node2vec       & LINE           & node2vec       & LINE           & doc2vec        & sentence2vec   & doc2vec        & sentence2vec   \\ \hline
%% \multirow{4}{*}{\begin{tabular}[c]{@{}c@{}}Hit@10 (\%) \\  in LP-Raw\end{tabular}}   & KG-O                   & 30.10          & \textbf{30.10} & 30.15          & 30.15          & \textbf{30.10} & \textbf{30.10} & 30.15          & 30.15          \\ 
%%                                                                                           & KG-I                   & 29.34          & 29.59          & 30.75          & 30.63          & 29.15          & 29.24          & \textbf{30.67} & 30.67          \\ 
%%                                                                                           & KG-P                   & \textbf{30.23} & 29.99          & \textbf{30.89} & \textbf{30.81} & 29.37          & 29.17          & 30.60          & \textbf{30.68} \\ 
%%                                                                                           & concat-O               & 27.12          & 28.21          & 28.71          & 28.32          & 28.10          & 29.04          & 30.21          & 29.01          \\ \hline\hline
\multirow{4}{*}{\begin{tabular}[c]{@{}c@{}}Hit@10 (\%) \\  in LP-Filtered\end{tabular}} & KG-O                   & 43.14          & 43.14          & 43.86          & 43.86          & \textbf{43.14} & \textbf{43.14} & 43.86          & 43.86          \\ 
                                                                                          & KG-I                   & 42.25          & 43.00          & 44.31          & 44.56          & 41.86          & 42.05          & 42.31          & \textbf{44.58} \\ 
                                                                                          & KG-P                   & \textbf{43.66} & \textbf{43.52} & \textbf{44.72} & \textbf{44.67} & 41.99          & 42.21          & \textbf{44.26} & 44.47          \\ 
                                                                                          & concat-O               & 36.99          & 37.47          & 38.32          & 38.45          & 40.17          & 40.07          & 40.79          & 37.83          \\ \hline \hline
\multirow{4}{*}{\begin{tabular}[c]{@{}c@{}}Accuracy (\%)\\ in TC\end{tabular}}            & KG-O                   & 76.56          & 76.56          & 78.29          & 78.29          & \textbf{76.56} & \textbf{76.56} & 78.29          & 78.29          \\ 
                                                                                          & KG-I                   & 76.70          & 76.86          & 78.54          & 78.80          & 76.06          & 76.42          & 78.63          & \textbf{78.61} \\ 
                                                                                          & KG-P                   & \textbf{77.13} & \textbf{77.09} & \textbf{78.96} & \textbf{79.11} & 76.17          & 76.21          & \textbf{78.70} & 78.60          \\ 
                                                                                          & concat-O               & 71.97          & 73.23          & 71.82          & 70.75          & 71.41          & 71.32          & 72.15          & 69.91          \\ \hline
\end{tabular}
\end{table}

%\vspace{-0.1in}
\subsection{A Large-Scale Dataset}\label{sec:experiment_large_dataset}
\begin{table}[]
\caption{Specifications for the large-scale dataset.}
\label{tbl:dataset_large}
\centering
\begin{tabular}{lllll}
\hline
\#ent. & \#scenario & \#category  & \#rel. & \#train \\ \hline
17.37M & 182K    & 8.96K  & 5.18K  & 60.65M  \\ \hline\hline
\#item & \#value & \#user & \#edge\_click & \#edge\_purchase \\ \hline
9.14M  & 8.04M & 482M & 7,952M & 144M \\\hline                      
\end{tabular}
%\vspace{-0.5cm}
\end{table}
In this section, we apply \model\ to the KG/BG embeddings generated from an Alibaba Taobao's large-scale dataset\footnote{The details of the Alibaba Taobao's dataset are deferred to Appendix \ref{appendix:data_detail}.}, whose statistics are summarized in Table \ref{tbl:dataset_large}. Considering the computational efficiency, TransE is used to get the KG embeddings on a knowledge database established by Alibaba Taobao.  As with the BG embeddings, we run GraphSAGE on a graph constructed in terms of users' behaviors, e.g., two items are connected if a certain number of customers bought them simultaneously over the past months. GraphSAGE is a representative work for graph neural network (GNN) and has achieved good performances for large datasets. %% \footnote{We do not use node2vec and LINE on the large dataset since they are not scalable. On the other hand, we observe that GraphSAGE achieves a quite lower accuracy than other methods on the two small datasets, so we do not report the results.}.
The dimension of KG embedding and the dimension of the BG embedding are $d_{\bw} = 64$, $d_{\bz} = 128$ as the online setting of Alibaba Taobao. We take the recommendation task for evaluation. Specifically, each customer has a set of trigger items from his/her historical behaviors including clicks, purchases, add-to-preferences and add-to-carts. These trigger items are then used to retrieve (by FAISS \cite{JDH17}) more items based on the BG embeddings. We evaluate our method by counting the number of retrieved items that will be actually bought/clicked by the user in the following days.  %% there is no model like a classifier to extract useful signals from it --- it masks the signals from the BG embeddings in most cases.
Table \ref{tbl:recommendation-result} exhibits the hit recall rates of the BG-P and BG-O on the recommendation task. %% We  that \model\/-P does not perform as well as \model\/-O if we consider the hit recall at the granular level of item ids. In fact,

We check whether the retrieved items are of the same brand/category as those actually bought/clicked items in the following days. Combining these two granularities, we observe that the hit recall rates for BG-P are boosted by $1\%$-$3\%$ compared to BG-O, which is quite significant considering there are over $9$ million items. It validates that \model\/-P is able to incorporate useful KG information into the BG embedding for the item recommendation purpose.

Finally, for each concept/scenario, we use TransE to predict its top $50$ item categories based on KG-O and KG-P (see the detailed procedure as Section \ref{sec:LP_TC}). The result shows that KG-P can find more related items for the given concepts, as shown in Table \ref{tbl:retrieval_examples}. It indicates that by incorporating the BG information via \model\/, we can acquire novel knowledge that does not exist in the original KG.

% \usepackage{multirow}
\begin{table}[]
  \caption{Hit recall rates (\%) for item recommendation based on customer-specific trigger items. The recommended items are retrieved by finding the closest items to the trigger items using KG embeddings by \model\/-P and \model\/-O.}
  \label{tbl:recommendation-result}
  \centering
\begin{tabular}{c|c|cc|cc}
\hline
\multirow{2}{*}{Granularity} & \multirow{2}{*}{Hit @} & \multicolumn{2}{c|}{click} & \multicolumn{2}{c}{buy} \\ \cline{3-6} 
                          &        & \model\/-O    & \model\/-P    & \model\/-O   & \model\/-P   \\ \hline
%% \multirow{3}{*}{item}     & 10     & 8.28        & \textbf{8.34}         & \textbf{10.49}      & 10.46       \\ 
%%                           & 30     & \textbf{8.45}        & 8.60         & 10.65      & \textbf{10.72}       \\ 
%%                           & 50     & \textbf{8.88}        & 8.67         & \textbf{10.95}      & 10.85       \\ \hline
\multirow{3}{*}{brand}    & 10     & 15.97       & \textbf{16.14}        & 24.87      & \textbf{25.10}       \\ 
                          & 30     & 16.65       & \textbf{17.12}        & 25.70      & \textbf{26.57}       \\ 
                          & 50     & 17.26       & \textbf{17.90}        & 26.39      & \textbf{27.33}       \\ \hline
\multirow{3}{*}{category} & 10     & \textbf{27.46}       & 27.40        & 27.85      & \textbf{27.91}       \\ 
                          & 30     & 28.43       & \textbf{29.99}        & 28.50      & \textbf{29.45}       \\ 
                          & 50     & 29.58       & \textbf{32.88}        & 29.26      & \textbf{31.47}       \\ \hline
\end{tabular}
%\vspace{-0.5cm}
\end{table}

\begin{table}[]
  \caption{Examples in which \model\ acquire novel knowledge that does not exist in the KG.}
  \label{tbl:retrieval_examples}
  \centering
\begin{tabular}{l|l|l}
\hline
concept         & predicted categories using KG-O                                                                                                                              & predicted categories using KG-P                                                                                                                                                                                                                                                                                                                       \\ \hline
neuter clothing & jacket, homewear                                                                                                                                                               & \begin{tabular}[c]{@{}l@{}}Quick-drying T-shirt,  sports down jacket\\  toning pants, aerobics clothes, warm pants\end{tabular}                                                                                                                                                                                                                                         \\ \hline
sports training & None                                                                                                                                                                           & \begin{tabular}[c]{@{}l@{}}Quick-drying T-shirt, sports down jacket, \\ Yoga T-shirt, training shoes, aerobics clothes, \\ sports bottle\end{tabular}                                                                                                                                                                                                                       \\ \hline
household items & \begin{tabular}[c]{@{}l@{}}succulents, detergent, tissue box,\\ kitchen knife,  man's facial cleanser,\\ washing cup, yoga mat towel,\\health tea,  scented candle \end{tabular} & \begin{tabular}[c]{@{}l@{}}washing machine cover, spray, table, tape,\\ fish tank cleaning equipment, pen container, \\ digital piano, maker, wood sofa bath bucket, \\ composite bed, mosquito patch, storage rack, \\   storage box, pillow interior,  leather sofa,\\ needle, cotton swab, laundry ball, coffee cup,\\ desiccant,  trash bag, indoors shoes,\end{tabular} \\ \hline
\end{tabular}
\end{table}

%%%%%%%%%%%%%%%%%
%% Discussion %%
%%%%%%%%%%%%%%%%%
\section{Discussion} \label{sec:conclusion}
In this paper, we introduce \model\/, a Bayesian framework that can refine graph embeddings by integrating the information from the KG and BG sources. \model\ has been evaluated on a variety of experiments. It is shown to be able to improve the embeddings on multiple tasks by leveraging the information from the other side. \model\ can achieve superior or comparable performance with higher efficiency to the concatenation method (the baseline) for the node classification task, and can help in other tasks where the simple aggregation methods (e.g., concatenation) are not applicable. It is designed by bridging KG and BG via a Bayesian generative model, where the former is regarded as the prior while the latter is the observation.  %It is designed by mimicing the cognition process of human beings, and thus holds the promise of refining the BG embeddings or even learning novel knowledge.

Currently, only one BG is considered at a time in this work. In fact, \model\ can be easily extended to deal with multiple BGs. The integration of more than one BGs may further refine the KG, as their behavior-specific biases can be mutually canceled out. Besides, for the time being, \model\ works only for pre-trained KG/BG embeddings. It can be potentially extended so that the networks for the KG/BG embeddings are connected and jointly trained via this framework. In other words, \model\ can act as an interface that connects any KG embedding method with any BG embedding method for the end-to-end training. This makes the learning of the BG embedding supervised by the KG information. In turn, the learning of the KG embedding can be supplemented with instantiated samples in BG.  %To the best of our knowledge, \model\ is the first work that can flexibly and seamlessly connect the KG side with the BG side, . 

%%%%%%%%%%%%%%
%% Appendix %%
%%%%%%%%%%%%%%
\clearpage
\appendix

\begin{center}
  {\huge Appendix}
\end{center}

\section{Dataset Details}\label{appendix:data_detail}
The data of our experiments based on public datasets mainly includes FB15K237, pagelink network, descriptions of entities and labels of entities. Their sources are discussed below.

\subsection{Small datasets}
\begin{figure}[htbp]
  \centering
  \includegraphics[width=0.6\textwidth]{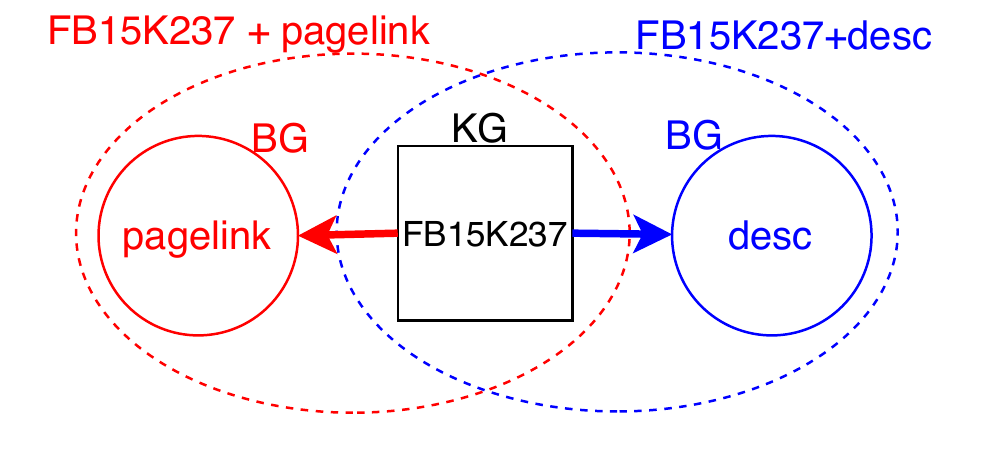}
  \caption{Illustration of KG and BG for the two small datasets.}
  \label{append_fig:diagram_two_small_datasets}
\end{figure}

The two small datasets share KG but differ in the BGs. Their relations are depicted in Figure \ref{append_fig:diagram_two_small_datasets}.

\textbf{Knowledge Graph} We use FB15k-237, a subset of Freebase, as the knowledge graph, which is also used in ConvE \cite{dettmers2018convolutional}. Different from the popular data set FB15k used in many knowledge graph representation researches, it does not include the inverse relations that may cause leakage from the training set to the validation set. FB15k-237 has 14,541 entities, 237 relations, 272,115 training triples, 20,466 test triples and 17,535 validation triples.

% \textbf{Knowledge Graph} We use FB15k, a subset of Wordnet, as the knowledge graph. It has 14951 entities, 1345 relations, 483142 training triples, 50000 validation triples and 59071 test triples. Each entity is a subject or an object in real life. Each triple is in the form of $<subject, relation, object>$.

\textbf{Pagelink Network} The pagelink network is a directed graph generated by ourselves. Since FB15k is a subset of Freebase, we first map the entities of FB15k to wikidata, that is a knowledge database to provide support for Wikipedia, Wikimedia Commons. according to the mapping data on the freebase database \cite{freebase2wikidata}. Then we use the pagelinks in English wikipedia to build the pagelink network. Since we could not get all the data, entities in the pagelink network are fewer than them in the knowledge graph. The pagelink network has 14,071 vertices and 1,065,412 edges in total.

\textbf{Descriptions of Entities} The descriptions used in our experiments are the same as DKRL \cite{xie2016representation}. It has 14,904 English descriptions of entities.

\textbf{Labels of Entities} In wikidata, the property 'instance of' is an isA relation which represents the class that the entity belongs to. Therefore, we use the property values of 'instance of' to represent the labels of entities used in the node classification task. At the same time, we also consider the problem of information leakage. In Freebase, the relation 'type/object/type' represents the type of an entity. To avoid that this relation may leak information to evaluation tasks, we check that the relation 'type/object/type' is not used in the triples of training set.

\subsection{Large dataset}
\textbf{Knowledge Graph of Alibaba Taobao} The knowledge graph of Alibaba Taobao items shows a tree structure. It contains four types of entities: items, categories items belong to, scenes of the categories, and the attribute values of the items. Therefore, there are three types of triples:
\begin{itemize}
    \item $<scene, cateOf, category>$,
    \item $<category, itemOf, item>$,
    \item $<item, property, property value>$.
\end{itemize}
Among the above three types of triplets, the first one is N-N mapping, the second
one is 1-N mapping and the third one is N-N mapping.

\textbf{Behavior Graph of Alibaba Taobao } The behavior graph of Alibaba Taobao is a bipartite graph that contains both user and item nodes. Interactions between users and items are CLICK or BUY which were sampled from a slicing window of 2 weeks (Dec. 27th, 2018 - Jan. 10th, 2019). The data of the first week was used for training. We used the trained model to recommend items for users with trigger items collected on Jan. 5th, 2019, and checked whether these recommended items were really clicked/bought in the following week.

\quad Each user has specific features describing their certain properties, e.g. age, gender, occupation, preference towards some category of items, the recently clicked items, and each item has features like price, category, brand, etc. Edges (interactions) have weights that decay with time. When learning the node embedding of the behavior graph, we use the edges between the user and the item as positive samples and randomly corrupted edges as negative samples. Node features are incorporated alone with edges in the training phase.

\section{Further details on functions}\label{appendix:func_detail}
To get embeddings of different data sets, we use several functions. The details of them are shown below.

\textbf{TransE} TransE is a typical knowledge graph representation method \cite{bordes2013translating}. It treats relations in knowledge graph as translating operators from head entities to tail entities, which is represented as
\begin{align}
    E(h,r,t)=||\textbf{h}+\textbf{r}-\textbf{t}||_{L1/L2}
\end{align}
In this work, we use the TransE API offered by \cite{han2018openke} to get embeddings of entities in knowledge graph.

\textbf{node2vec} Node2vec is a network representation framework \cite{grover2016node2vec}. It uses a biased random walk procedure to preserve the neighborhood information of the network in node representation. We believe the neighborhood information in the pagelink network can help characterize an entity, so we use it to generate vertex embeddings of pagelink network. In our experiment, we set the parameters as follows: the length of walk is 80, the number of walks is 10, the context size is 10.

\textbf{LINE} LINE is a network representation method \cite{tang2015line}. It preserves the first-order
and second-order proximities in a network. In this work, we use the LINE API offered by OpenKE to get entity embeddings in the pagelink network. In our experiment, we set the negative ratio is 5, and uses both the 1st-order and the 2nd-order proximity of graphs.

\textbf{doc2vec} Doc2vec is an unsupervised framework to get embeddings of given sentences or paragraphs \cite{le2014distributed}. Embeddings of documents are trained to predict the words according to its context in the documents. We use it to get entity embeddings based on entity descriptions. In our experiment, we use PV-DM (Distributed Memory Model of paragraph vectors) to get the embeddings of documents.

\textbf{sentence2vec} Sentence2vec is an unsupervised, C-BOW-inspired framework to get embeddings of sentences or paragraphs \cite{pgj2017unsup}. It has been proven to have a state-of-the-art performance in sentence similarity comparison task. Therefore, we use it to generate entity embeddings based on descriptions for the purpose of reconstruct the graph based on vertex similarity. In our experiment, we set the parameters as follows: the learning rate is 0.2, the update rate of learning rate is 100, the number of epochs is 5, the minimal number of word occurrences is 5, the minimal number of label occurrences is 0, the max length of word gram is 2.

\textbf{GraphSAGE} GraphSAGE is an inductive representation learning framework. Unlike transductive graph embedding frameworks that only generate embeddings for seen nodes, GraphSAGE leverages node attribute information to learn node embeddings in a generalized way and thus is capable of generating representations on unseen data. We use GraphSAGE to learn node embeddings on Alibaba Taobao's Behavior Graph.

\section{The selection of parameters for Algorithm \ref{algo:BEM-full}}\label{appendix:algo_parameters}
To understand how the tuning parameters influences the performance of \model\/-P, we apply Algorithm \ref{algo:BEM-full} to pre-trained FB15K237 embeddings (KG) obtained by TransE and pre-trained \textit{pagelink} embeddings obtained by node2vec. Each time we only change one parameter based on the default setup mentioned in Section \ref{sec:experiments}, i.e., $n_h = 500$, $n_B = 500$, $\text{learning rate} = 0.001$, $T \cdot n_B/N = 20$, $\lambda_1 = \lambda_2 = 1.0$. The associative results of link prediction and triplet classification are displayed in Table \ref{appendix_tbl:algo_parameters}. We can draw a few conclusions from such results:
\begin{itemize}
\item $n_h$, $n_B$ and the number of training steps $T$ affect the \model\/-P marginally. It indicates that \model\/-P does not require high model complexity for expressiveness and converges quickly.  
\item The learning rate is worth tuning as other gradient-based algorithms.
\item The most important parameters are $\lambda_1$ and $\lambda_2$. As explained in the last paragraph of Section \ref{sec:influence_model}, they balance the reconstruction term and the penalty term in Equation \eqref{eq:recons-indpt-pair} and Equation \eqref{eq:penalty-indpt-pair}. Tuning $\lambda_1$ and $\lambda_2$ based on a validation set might give significant boost in performance. But if the user wants to skip tuning, $\lambda_1 = 1$ and $\lambda_2 = 1$ can be the good starting point.
\end{itemize}
For the node classification task, we get similar results using the same dataset.

\begin{table}[ht]
  \caption{The results of link prediction and triplet classification for the TransE method on the FB15K237 dataset and the node2vec method on the \textit{pagelink} dataset, with varying tuning parameters. The default parameters are $n_h = 500$, $n_B = 500$, $\text{learning rate} = 0.001$, $T \cdot n_B/N = 20$, $\lambda_1 = \lambda_2 = 1.0$. Each row in the table only changes one parameter while keeping the others the same as default.}
  \label{appendix_tbl:algo_parameters}
  \centering
\begin{tabular}{c|cccc||cccc}
    
\hline
                  & \multicolumn{4}{c||}{Hit@10 (10\%) in LP-Filtered}                 & \multicolumn{4}{c}{Accuracy (\%) in TC}                          \\ \hline\hline
$n_h$   & 200            & 500            & 800            &                & 200            & 500            & 800            &                \\ 
evaluation result        & 43.38          & \textbf{43.66} & 43.41          &                & \textbf{77.25} & 77.13          & 77.19          &                \\ \hline
$n_B$             & 100            & 500            & 1000           &                & 100            & 500            & 1000           &                \\ 
evaluation result        & \textbf{43.77} & 43.66          & 42.63          &                & \textbf{77.40} & 77.13          & 77.28          &                \\ \hline
learning rate     & 0.0001         & 0.001          & 0.005          & 0.01           & 0.0001         & 0.001          & 0.005          & 0.01           \\ 
evaluation result        & 42.53          & 43.66          & 44.95          & \textbf{45.10} & 76.59          & 77.13          & \textbf{77.89} & 77.74          \\ \hline
$T \cdot n_B/N$ & 10             & 20             & 50             & 100            & 10             & 20             & 50             & 100            \\ 
evaluation result        & 43.77          & 43.66          & \textbf{44.03} & 43.83          & 76.55          & 77.13          & \textbf{77.39} & 76.97          \\ \hline
$\lambda_1$       & 0.01           & 0.1            & 1              & 5              & 0.01           & 0.1            & 1              & 5              \\ 
evaluation result        & 44.91          & \textbf{45.82} & 43.66          & 31.90          & 78.21          & \textbf{79.33} & 77.13          & 71.08          \\ \hline
$\lambda_2$       & 0.01           & 0.1           & 1              & 5              & 0.01          & 0.1            & 1              & 5              \\ 
evaluation result        & 41.59          & 41.74          & 43.66          & \textbf{44.35} & 76.11          & 76.61          & 77.13          & \textbf{77.41} \\ \hline
\end{tabular}
\end{table}

\section{More results of the node classification task on the FB15K237 dataset with two associative BGs}\label{appendix:results}

It is unfair to compare the \model\/-refined embedding to the concatenated embedding directly, since the latter is longer than the former. In our case, the length of the concatenated embedding is $2$ ($0.5$) times longer than that of the KG (BG) embedding. Thus, the classifier for the concatenation has $2$ ($0.5$) more parameters than that of the KG (BG) embedding. To get a fair comparison, we project the concatenated embedding into $\R^{d_{\bw}}$ ($\R^{d_{\bz}}$) using a random Gaussian projection matrix, which can nearly preserve the distances between nodes. 

Table \ref{appendix-tbl:cate-classification-pagelink} and \ref{appendix-tbl:cate-classification-desc} illustrate the results of \model\ with its variations on the node classification results. Four implications can be drawn by looking at the table in different ways. First, we observe consistent improvements of \model-P over \model\/-O through all settings. The classification accuracies on the BG (KG) embedding are boosted by about $2\%$-$10\%$ with \model\/-P. As for the concatenation version, the concat-O vector is expected to work better than embeddings by \model\ if the classifier is expressive enough --- there might be loss of information during the procedure of the \model\ integration. However, it turns out that concat-P outperforms concat-O. It indicates that \model\/-P does not lose information related to the classification task, and is able to make the embeddings into a better shape for the classification task. Second, for a fair comparison in terms of the dimension, we use Gaussian random projections (repeated for $10$ times) to project the concatenated embedding to $\R^{50}$ and $\R^{100}$, respectively. KG-P is superior to the projections of concat-O (for both $\R^{50}$ and $\R^{100}$), and is even comparable to concat-O. From the perspective of dimension reduction, this result suggests that \model\/-P can preserve the majority of information for KG. On the other hand, considering the goal of preserving the topological structure, \model\/-P is unlikely to boost the performance of low-quality BG-O to the level of concat-O. Third, we note the projections of concat-P loses marginal power during the dimension reduction, and are more robust than the projections of concat-O. It indicates that the \model\/-P representation is less noisy than the original embeddings. Finally, as we expect, \model-P outperforms the \model-I where the former accounts for the pairwise interactions. Such key information is crucial for the learning of both the KG and BG embeddings.

% \usepackage{multirow}
\begin{table}[]
  \caption{The node classification accuracy (\%) on the FB15K237-\textit{pagelink} dataset, using the BG/KG embeddings refined by \model\/. Here concat$\shortrightarrow\R^{50}$/concat$\shortrightarrow\R^{100}$ refers to the projection of concat into $\R^{50}$/$\R^{100}$, and concat refers to the concatenated embedding itself. The numbers in the brackets of concat$\shortrightarrow\R^{50}$/$\R^{100}$ are the standard errors across $10$ random projections.}
  \label{appendix-tbl:cate-classification-pagelink}
    \centering
\begin{tabular}{c|c|ccccc}
\hline
\multicolumn{2}{c|}{}                                          & \multicolumn{5}{c}{node2vec}                                                                         \\ \hline
                      & \model\ & KG             & BG             & concat$\shortrightarrow\R^{50}$ & concat$\shortrightarrow\R^{100}$ & concat    \\ \hline
\multirow{3}{*}{TransE} & O                                     & 85.59          & 75.12          & 82.86 (0.93)            & 87.58 (0.27)             & 89.39          \\
                        & I                                     & 85.51          & 82.56          & 83.96 (0.35)            & 85.39 (0.16)             & 85.97          \\
                        & P                                     & \textbf{88.89} & \textbf{86.32} & \textbf{88.71 (0.20)}   & \textbf{89.27 (0.16)}    & \textbf{90.29} \\ \hline
\multirow{3}{*}{TransD} & O                                     & 86.06          & 75.12          & 82.37 (0.82)            & 86.24 (0.26)             & 89.18          \\
                        & I                                     & 83.73          & 78.86          & 81.67 (0.80)            & 83.83 (0.28)             & 84.16          \\
                        & P                                     & \textbf{88.60} & \textbf{85.39} & \textbf{87.83 (0.26)}   & \textbf{88.92 (0.27)}    & \textbf{89.90} \\ \hline\hline
\multicolumn{2}{c|}{}                                          & \multicolumn{5}{c}{LINE}                                                                             \\ \hline
                      & \model                                   & KG             & BG             & concat$\shortrightarrow\R^{50}$ & concat$\shortrightarrow\R^{100}$ & concat    \\ \hline
\multirow{3}{*}{TransE} & O                                     & 85.59          & 77.57          & 79.49 (1.14)            & 86.54 (0.40)             & 89.44          \\ 
                        & I                                     & 86.35          & 85.44          & 85.65 (0.34)            & 86.63 (0.14)             & 87.05          \\ 
                        & P                                     & \textbf{88.21} & \textbf{86.27} & \textbf{88.21 (0.48)}   & \textbf{89.12 (0.15)}    & \textbf{90.01} \\ \hline
\multirow{3}{*}{TransD} & O                                     & 86.06          & 77.57          & 77.51  (1.16)           & 85.80 (1.03)             & 89.00         \\ 
                        & I                                     & 86.58          & 85.10          & 85.36 (0.35)            & 86.40 (0.18)             & 86.69          \\ 
                        & P                                     & \textbf{88.70} & \textbf{85.30} & \textbf{87.95 (0.44)}   & \textbf{88.82 (0.15)}    & \textbf{89.73} \\ \hline
\end{tabular}
\end{table}

% \usepackage{multirow}
\begin{table}[]
  \caption{The node classification accuracy (\%) on the FB15K237-\textit{desc} dataset, using the BG/KG embeddings refined by \model\/. Here concat$\shortrightarrow\R^{50}$/concat$\shortrightarrow\R^{100}$ refers to the projection of concat into $\R^{50}$/$\R^{100}$, and concat refers to the concatenated embedding itself. The numbers in the brackets of concat$\shortrightarrow\R^{50}$/$\R^{100}$ are the standard errors across $10$ random projections.}
  \label{appendix-tbl:cate-classification-desc}
  \centering
\begin{tabular}{c|c|ccccc}
\hline
\multicolumn{2}{c|}{}                                          & \multicolumn{5}{c}{doc2vec}                                                                          \\ \hline
                        & \model\ & KG             & BG             & concat$\shortrightarrow\R^{50}$ & concat$\shortrightarrow\R^{100}$ & concat    \\ \hline
\multirow{3}{*}{TransE} & O                                     & 85.32          & 75.62          & 83.87 (0.63)            & 86.77 (0.23)             & \textbf{87.92} \\ 
                        & I                                     & 86.19          & 81.50          & 85.32 (0.48)            & 86.10 (0.08)             & 86.41          \\ 
                        & P                                     & \textbf{87.68} & \textbf{81.52} & \textbf{86.40 (0.21)}   & \textbf{87.78 (0.21)}    & 87.86          \\ \hline
\multirow{3}{*}{TransD} & O                                     & 85.83          & 75.62          & 83.19 (0.62)            & 86.57 (0.35)             & 88.07          \\ 
                        & I                                     & 86.75          & 81.44          & 85.48  (0.73)           & 86.52 (0.13)             & 86.85          \\ 
                        & P                                     & \textbf{87.34} & \textbf{82.24} & \textbf{86.31 (0.43)}   & \textbf{87.57 (0.19)}    & \textbf{88.15} \\ \hline\hline
\multicolumn{2}{c|}{}                                          & \multicolumn{5}{c}{sentence2vec}                                                                     \\ \hline
                        & \model\                                   & KG             & BG             & concat$\shortrightarrow\R^{50}$ & concat$\shortrightarrow\R^{100}$ & concat    \\ \hline
\multirow{3}{*}{TransE} & O                                     & 85.32          & 83.42          & 86.67 (0.39)            & 87.58 (0.24)             & 88.43          \\ 
                        & I                                     & 87.61          & 85.18          & 86.95 (0.31)            & 87.70 (0.14)             & 88.07          \\ 
                        & P                                     & \textbf{88.05} & \textbf{85.82} & \textbf{87.61 (0.16)}   & \textbf{88.36 (0.24)}    & \textbf{88.57} \\ \hline\hline
\multirow{3}{*}{TransD} & O                                     & 85.83          & 83.42          & 85.69 (0.32)            & 87.83 (0.16)             & 88.52          \\ 
                        & I                                     & 87.96          & 84.97          & 86.89 (0.4)             & 87.91 (0.24)             & 88.07          \\ 
                        & P                                     & \textbf{88.36} & \textbf{86.12} & \textbf{87.40 (0.20)}   & \textbf{89.59 (0.18)}    & \textbf{88.86} \\ \hline
\end{tabular}
\end{table}

\bibliographystyle{plain}
\bibliography{Bayes_Emb}
\end{document}